\documentclass{article} %
\usepackage{iclr2024_conference,times}

\usepackage{amsmath,amsfonts,bm}

\def\eqref#1{equation~\ref{#1}}

\def\floor#1{\lfloor #1 \rfloor}
\def\1{\bm{1}}

\def \btheta{{\boldsymbol{\theta}}}
\def \bphi{{\boldsymbol{\phi}}}
\def \bfeta{{\boldsymbol{\eta}}}
\def \diag{{\operatorname{diag}}}
\def \bnu{{\boldsymbol{\mu}}}
\def \bsigma{{\boldsymbol{\sigma}}}
\def \bLambda{{\boldsymbol{\Lambda}}}
\def \bOmega{{\boldsymbol{\Omega}}}

\def\rp{{\textnormal{p}}}
\def\rq{{\textnormal{q}}}

\def\vf{{\bm{f}}}
\def\vg{{\bm{g}}}
\def\vh{{\bm{h}}}

\def\vq{{\bm{q}}}

\def\vv{{\bm{v}}}

\def\vx{{\bm{x}}}
\def\vy{{\bm{y}}}
\def\vz{{\bm{z}}}

\DeclareMathAlphabet{\mathsfit}{\encodingdefault}{\sfdefault}{m}{sl}
\SetMathAlphabet{\mathsfit}{bold}{\encodingdefault}{\sfdefault}{bx}{n}

\newcommand{\softmax}{\mathrm{softmax}}

\def\rP{{\textnormal{P}}}

\usepackage{hyperref}
\usepackage{url}
\usepackage{dsfont}
\usepackage{mathtools}
\usepackage{algorithm}
\usepackage{algpseudocode}
\usepackage{booktabs}
\usepackage{enumitem}
\usepackage{caption}[skip=0pt]
\usepackage{nicefrac}

\title{It HAS to be Subjective: Human Annotator Simulation via Zero-shot Density Estimation}

\author{Wen Wu$^{*,1}$, Wenlin Chen$^{*,1,2}$\thanks{$^{*}$Equal contribution.},  Chao Zhang$^{3}$, Philip C. Woodland$^{1}$  \\
$^{1}$ Department of Engineering, University of Cambridge, Cambridge, UK\\
$^{2}$ Department of Empirical Inference, Max Planck Institute for Intelligent Systems, T\"{u}bingen, Germany\\
$^{3}$Department of Electronic Engineering, Tsinghua University, Beijing, China\\
\texttt{\{ww368, wc337, pcw\}@eng.cam.ac.uk}, \texttt{cz277@tsinghua.edu.cn} \\
}

\iclrfinalcopy %
\begin{document}

\maketitle

\begin{abstract}
Human annotator simulation (HAS) serves as a cost-effective substitute for human evaluation such as data annotation and system assessment. Human perception and behaviour during human evaluation exhibit inherent variability due to diverse cognitive processes and subjective interpretations, which should be taken into account in modelling to better mimic the way people perceive and interact with the world. This paper introduces a novel meta-learning framework that treats HAS as a zero-shot density estimation problem, which incorporates human variability and allows for the efficient generation of human-like annotations for unlabelled test inputs. Under this framework, we propose two new model classes, conditional integer flows and conditional softmax flows, to account for ordinal and categorical annotations, respectively. The proposed method is evaluated on three real-world human evaluation tasks and shows superior capability and efficiency to predict the aggregated behaviours of human annotators, match the distribution of human annotations, and simulate the inter-annotator disagreements.

\end{abstract}

\section{Introduction}
Human evaluation is fundamental to machine learning research, guiding processes such as data annotation and model assessment, which for instance include perceptual quality evaluation of synthesized speech, text, and image~\citep{ma2015perceptual,patton2016automos,fu18c_interspeech,talebi2018nima,lo19_interspeech,borade2021automated,ramesh2022automated}, annotation generation for weak supervision~\citep{NIPS2016_6709e8d6,wu2022learning}, and model optimization based on human preference~\citep{schatzmann2007agenda,asri16_interspeech,gur2018user,ruiz2018learning,shi-etal-2019-build,lin-etal-2021-domain}. 
Collecting human annotations or evaluations often requires substantial resources and may expose human annotators to distressing and harmful content in sensitive tasks (\textit{e.g.}, toxic speech detection, suicidal risk prediction, and depression detection). This inspires the exploration of human annotator simulation (HAS) as a scalable and cost-effective alternative, which facilitates large-scale dataset evaluation, benchmarking, and system comparisons.

Variability is a unique aspect of real-world human evaluation, since individual variations in cognitive biases, cultural backgrounds, and personal experiences~\citep{hirschberg2003experiments,wiebe2004learning,haselton2015evolution} can lead to variability in human interpretation~\citep{MSP-podcast,mathew2021hatexplain,maniati22_interspeech}. HAS aims to incorporate the variability present in human evaluation rather than solely relying on majority opinions, which mitigates potential biases and over-representation in scenarios where dominant opinions could potentially overshadow minority viewpoints~\citep{dixon2018measuring,hutchinson-etal-2020-social}, thus promoting fairness and inclusivity.

In this work, we investigate HAS for the automatic generation of human-like annotations that take into account the variability in human evaluation. A novel meta-learning framework that treats HAS as a zero-shot density estimation problem is introduced, which allows for the efficient generation of human-like annotations for unlabelled test inputs. Under this framework, two new model classes, conditional integer flows and conditional softmax flows, are proposed to account for ordinal and categorical annotations respectively, which are common types of annotations in human evaluation tasks. The proposed method shows superior capability and efficiency to predict the aggregated behaviours of human annotators, match the distribution of human annotations, and simulate the level of inter-annotator agreement on three 
real-world human evaluation tasks: emotion recognition, toxic speech detection, and speech quality assessment. 

\section{Human Annotator Simulation (HAS)}
\label{sec: SALS}

\subsection{The Variability in Human Evaluation is Valuable}
Each individual's perception of the world is unique and influenced by their physical state and cognitive biases, which leads to diverse and subjective interpretations (see Appendix~\ref{apdx: variability in HAS} for more detail). Such subjectivity can be manifest in various tasks such as  emotion recognition~\citep{hirschberg2003experiments,mihalcea2006corpus}, perceptual quality assessment~\citep{wiebe2004learning,5404314,zen2016assessing}, and user experience evaluation~\citep{zen2016assessing}. It has been argued that achieving a deterministic ``ground truth'' in subjective tasks like human evaluation is not feasible, nor essential ~\citep{alm2011subjective,wu2022estimating}. Therefore, we advocate for methodologies that focus on modelling annotators' subjective interpretations, rather than seek to reduce the variability in annotations: instead of only predicting the majority opinion, it is important to account for the human perception variability when designing a human annotator simulator. The following are three examples that demonstrate the importance of modelling variability in HAS:

\textbf{Revealing data ambiguity.} 
Incorporating the variability in human perception empowers HAS to reveal potential ambiguity or complexity in data, providing valuable insights for further analysis.

\textbf{Mitigating bias and over-representation.} 
Incorporating the variability in human judgements prevents HAS from being biased towards a certain perspective and ignoring minority viewpoints, leading to a more inclusive representation of opinions where all viewpoints are given due consideration.

\textbf{Improving model alignment.}
Optimization based on human feedback has led to superior performance on tasks such as text generation~\citep{christiano2017deep,ouyang2022training,rafailov2023direct}, which aligns the behaviour of language models with human preferences. HAS could be helpful in this task, as it is an efficient and cost-effective alternative to generating human feedback. 

\subsection{Problem Formulation and Related Work}
HAS involves modelling  a dataset $\mathcal{D}=\{(\vx_i,\bfeta_i^{(1)},\cdots,\bfeta_i^{(M_i)})\}_{i=1}^N$ from human evaluation, where each data point is a tuple of an input $\vx_i$ and its corresponding labels $\bfeta_i^{(1)},\cdots,\bfeta_i^{(M_i)}$ provided by $M_i$ independent human annotators. Note that different inputs may be labelled by different sets of annotators. HAS aims to model the conditional annotation distribution $\rp(\bfeta|\vx)$ in order to simulate human-like annotations $\bfeta_*^{(1)},\cdots,\bfeta_*^{(M_*)}$ for any unseen input $\vx_*$ in a way that reflects how it would be labelled by human annotators. Prior work mainly investigated two approaches to this problem.

The first approach uses a single proxy variable $\bfeta_i$ (\textit{e.g.}, majority vote or average score) to summarize all annotations for each input $\vx_i$~\citep{Kim_2013,djuric2015hate,patton2016automos,Poria2017}. This creates a proxy dataset $\mathcal{D}'=\{(\vx_i,\bfeta_i)\}_{i=1}^N$ and converts HAS into a supervised learning problem, which is usually solved by fitting a discriminative model to estimate the conditional distribution for the proxy variable. This approach assumes that each input $\vx_i$ has only one ground-truth label $\bfeta_i$, and thus the conditional distribution only quantifies the uncertainty due to noisy observation and lack of training examples. 
Clearly, modelling a single proxy variable as in this approach fails to take into account the subjectivity and diversity in human behaviour and perception and thus will result in an underestimate of variability in the simulated annotations. Other work incorporated the variance of human annotations into the proxy variable~\citep{deng2012confidence,prabhakaran2015statistical,plank2014learning,dang2017investigation,han2017hard,leng2021mbnet}. However, all these approaches still focus on obtaining the ``correct'' label (\textit{e.g.}, aiming for improved prediction accuracy) and minimizing the discrepancy among annotators (\textit{e.g.}, reducing ``noise'' in annotations) rather than embracing inter-annotator disagreements.

The second approach explicitly models the behaviour of different annotators using different individual models in an ensemble or different heads in a single model \citep{Fayek_2016,chou2019every,davani2022dealing}. However, this approach is computationally feasible only when the number of annotators is relatively small and when a sufficient quantity of annotation is available for each annotator, which can then not be applied to large crowd-sourced datasets, such as \cite{MSP-podcast,mathew2021hatexplain,maniati22_interspeech}, 
which are common in real-world applications.

\begin{figure}[tb]
    \centering
    \includegraphics[width=\textwidth]{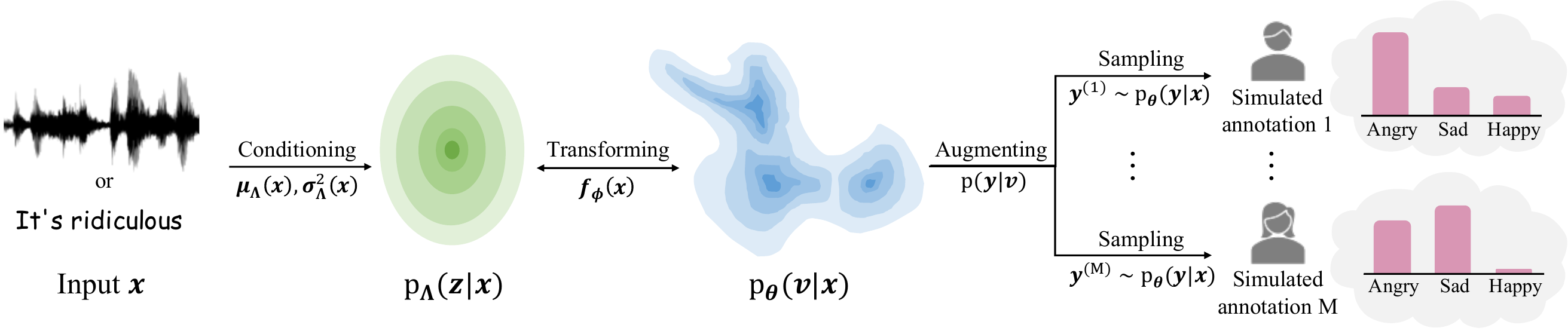}
    \caption{Diagram for the proposed zero-shot human annotator simulation framework.}
    \label{fig:illustrative_figure}
\end{figure}

\section{Human Annotator Simulation via Zero-shot Density Estimation}
\label{sec: CNF}

\subsection{A Meta-learning Framework for Zero-shot Human Annotator Simulation}
We propose a novel meta-learning framework for HAS to address the issues in prior work, where the collection of all human annotations for each input $\vx_i$ is viewed as a dataset $\mathcal{D}_i=\{\bfeta_i^{(m)}\}_{m=1}^{M_i}$. This framework transforms HAS into a density estimation problem for the human annotation of each input $\vx_i$ given samples from $\mathcal{D}_i$. We propose to meta-learn a density estimator across all datasets $\mathcal{D}_\text{meta}=\{\mathcal{D}_i\}_{i=1}^N$, in order to leverage knowledge about the agreements and disagreements among different human annotators across different examples. This formulates a \emph{zero-shot density estimation} problem, since there is no human annotation available for test-time adaptation except for a ``descriptor'' (\textit{i.e.}, the test input) $\vx_*$. 
In other words, the meta-learned density estimator should allow efficient sampling of human-like annotations and their likelihoods evaluations for any unseen input $\vx_*$ without access to any samples for the ground-truth human annotations of $\vx_*$.

In this work, the meta-learning framework is realized using a latent variable model\footnote{For clarity, we use different notations for human annotations $\bfeta$ and model outputs $\vy$.}:
\begin{align}
    \rp_{\btheta}(\vy|\vx)=\int \rp(\vy|\vv)\rp_{\bphi}(\vv|\vz)\rp_{\bLambda}(\vz|\vx)d\vv d\vz,\label{eq:formulation}
\end{align}
where the conditional prior $\rp_{\bLambda}(\vz|\vx)$ learns to summarize useful information about $\vx$ and encode the possible disagreements over $\vx$ among different human annotators, which is helpful for the likelihood $\rp_{\bphi}(\vy|\vz)=\int\rp(\vy|\vv)\rp_{\bphi}(\vv|\vz)d\vv$ to simulate human-like annotations.

Figure \ref{fig:illustrative_figure} illustrates the proposed framework workflow. Specifically, the conditional prior is modelled by a conditional factorized Gaussian distribution $\rp_\bLambda(\vz|\vx)=\mathcal{N}(\vz|\bnu_\bLambda(\vx), \diag(\bsigma_\bLambda^2(\vx)))$ whose mean $\bnu_\bLambda(\vx)$ and variance $\bsigma_\bLambda^2(\vx)$ are parameterized by a neural network with parameters $\bLambda$. 
The intermediate variable $\vv$ is obtained by a deterministic invertible transformation $\rp_{\bphi}(\vv|\vz)=\delta(\vv-\vf_{\bphi}(\vz))$, where $\vf_{\bphi}(\vz)$ is parameterized by an invertible neural network with parameters $\bphi$, and $\delta(\cdot)$ is the multivariate Dirac delta function. This results in a conditional normalizing flow (CNF):
\begin{align}
    \rp_\btheta(\vv|\vx)
    =\int \delta(\vv-\vf_{\bphi}(\vz))\rp_{\bLambda}(\vz|\vx)d\vz
    =\rp_\bLambda\left.\left(\vf^{-1}_\bphi(\vv)\right|\vx\right) \left|\operatorname{det}\left(\frac{\partial \vf^{-1}_\bphi(\vv)}{\partial \vv} \right) \right|,
    \label{eq: cnf p(y|x)}
\end{align}
where $\operatorname{det}(\cdot)$ denotes the determinant operator, ${\partial \vf^{-1}_\bphi(\vv)}/{\partial \vv}$ denotes the Jacobin matrix of $\vf^{-1}_\bphi(\vv)$, and $\btheta\coloneqq\{\bphi,\bLambda\}$ denotes all parameters in this base CNF. This modelling choice has the advantage of having tractable marginal likelihood as in Eqn.~(\ref{eq: cnf p(y|x)}) while not restricting the intermediate variable $\vv$ to a specific type of distribution as in previous methods, \textit{e.g.}, Gaussian~\citep{han2017hard} and Student-t ~\citep{wu-etal-2023-estimating} distributions, thus offering enhanced tractability, flexibility and generality. In addition, samples can be efficiently drawn from this model by first drawing $\vz\sim \rp_{\bLambda}(\vz|\vx)$ from the conditional prior and then computing the deterministic flow transformation $\vv=\vf_{\bphi}(\vz)$.

Finally, the output variable $\vy$ is obtained by augmenting the intermediate variable $\vv$ using the transformation $\rp(\vy|\vv)$, in order to accommodate different types of annotation. For continuous annotations, the identity transformation $\rp(\vy|\vv)=\delta(\vy-\vv)$ is used, which exactly recovers the base CNF model. However, real-world human evaluation tasks often involve discrete annotations that are either ordinal or categorical. In the following sections, two new model classes with meta-learning objectives are introduced to accommodate these annotation types.

\subsection{Meta-learning Conditional Integer Flows for Ordinal Annotations}\label{ssec:i-cnf}
\textbf{Modelling.} Discrete ordinal annotations are often used in K-point rating systems, where the ratings are integer-valued with a clear ordering. We propose a new class of models called conditional integer flows (I-CNFs), which augment the base CNFs by quantizing the continuous intermediate variable $v$ to its nearest integer by using a rounding transformation $\rp(y|v)=\mathbb{I}(y-\nicefrac{1}{2}<v\leq y+\nicefrac{1}{2})$, where $\mathbb{I}(\cdot)$ is the indicator function.
Let $o$ be an ordinal variable that represents the ordinal human rating for an input $\vx$. The marginal likelihood of I-CNF is given by
\begin{align}\label{eq:exact mlkl i-cnf}
    \rp_{\btheta}(o=y|\vx)=\int_{-\infty}^{\infty} \mathbb{I}(y-\nicefrac{1}{2}<v\leq y+\nicefrac{1}{2})\rp_{\btheta}(v|\vx)dv=\int_{y-\nicefrac{1}{2}}^{y+\nicefrac{1}{2}}\rp_{\btheta}(v|\vx)dv,
\end{align}
where $\rp_{\btheta}(v|\vx)$ is the marginal likelihood of the base CNF defined in Eqn.~(\ref{eq: cnf p(y|x)}). Since the marginal likelihood of I-CNF given in Eqn.~(\ref{eq:exact mlkl i-cnf}) is analytically intractable due to the rounding transformation, we propose to approximate it using numerical integration. In practice, the rectangular rule is found to work well in terms of both performance and efficiency in this setting, where the density of $\rp_{\btheta}(v|\vx)$ within the interval $v\in(y-\nicefrac{1}{2},y+\nicefrac{1}{2}]$ is approximated by the midpoint density value:
\begin{align}\label{eq:i-cnf approx}
    \small
    \int_{y-\nicefrac{1}{2}}^{y+\nicefrac{1}{2}}\rp_{\btheta}(v|\vx)dv
    \approx \left(\left(y+\frac{1}{2}\right)-\left(y-\frac{1}{2}\right)\right)\cdot\rp_{\btheta}\left(\left.\frac{(y-\nicefrac{1}{2})+(y+\nicefrac{1}{2})}{2}\right|\vx\right)
    =\rp_{\btheta}(y|\vx).
\end{align}
This means that Eqn.~(\ref{eq: cnf p(y|x)}) can be used as a proxy to evaluate the likelihood of I-CNF.

\textbf{Meta-learning.} Using the numerical approximation given in Eqn.~(\ref{eq:i-cnf approx}), the loss $\mathcal{L}(\btheta;\mathcal{D}_i,\vx_i)$ for I-CNF on a single dataset $\mathcal{D}_i$ can be defined as the average negative log marginal likelihood of Eqn.~(\ref{eq: cnf p(y|x)}) evaluated on the human annotations $\mathcal{D}_i=\{\eta_i^{(m)}\}_{m=1}^{M_i}$ given the corresponding input $\vx_i$:
\begin{equation}
\begin{split}
    \mathcal{L}(\btheta; \mathcal{D}_i,\vx_i) 
    &= - \frac{1}{M_i}\sum_{m=1}^{M_i} \left( \log\rp_\bLambda\left.\left(f^{-1}_\bphi(\eta_i^{(m)})\right|\vx_i\right) +\log\left|\operatorname{det}\left(\frac{\partial f^{-1}_\bphi(\eta_i^{(m)})}{\partial \eta_i^{(m)}} \right) \right| \right).
\end{split}
\end{equation}
Following the episodic training scheme \citep{vinyals2016matching,snell2017prototypical,chen2023metalearning}, we treat density estimation on each dataset as a learning problem and randomly sample a subset of such learning problems to train on at each step during meta-training. This results in a meta-learning objective across all datasets $\mathcal{D}_\text{meta}$:
\begin{align}
    \mathcal{L}_\text{meta}(\btheta;\mathcal{D}_\text{meta},\{\vx_i\}_{i=1}^N)=\mathbb{E}_{D_i\sim\rp(\mathcal{D})}[\mathcal{L}(\btheta;\mathcal{D}_i,\vx_i)],
\end{align}
where $\rp(\mathcal{D})$ denotes the uniform distribution over the datasets in $\mathcal{D}_\text{meta}$. Intuitively, this objective maps all human annotation to the latent space of their corresponding input by the inverse flow transformation $z_i^{(m)}=f^{-1}_\bphi(\eta_i^{(m)})$ during meta-training, which helps the model to build a diverse latent representation that captures the variability in human annotations across different inputs. 
At test time, the I-CNF can simulate human-like annotations for an unseen, unlabelled input $\vx_*$ by first drawing $v_*^{(m)}\sim \rp_{\btheta}(v|\vx_*)$ from the base CNF then applying the rounding function $y_*^{(m)}=\floor{v_*^{(m)}}$, for $m=1,\cdots,M_*$, where $M_*$ denotes the number of annotations to be simulated.

\subsection{Meta-learning Conditional Softmax Flows for Categorical Annotations}
\textbf{Modelling.} To account for non-ordinal categorical annotations (\textit{e.g.}, emotion categories), we propose a new class of models called conditional softmax flows (S-CNFs), which augments the base CNFs by applying the softmax function $\rp(\vy |\vv) = \delta\left(\vy-\softmax(\vv)\right)$ to transform the continuous intermediate variable $\vv$ into categorical probabilities $\vy$.
Let $c$ be a categorical variable with probability $\rP(c=k|\vy)=\vy_k~(k=1,\cdots,K)$ that represents the categorical human annotation for an input $\vx$, with $\rP(c=k|\vv)=\int\vy_k\delta\left(\vy-\softmax(\vv)\right)d\vy=\softmax(\vv)_k$. The marginal likelihood of S-CNF is given by
\begin{equation}\label{eq: S-CNF-mlkl}
    \rP_\btheta(c=k|\vx) 
    = \int \rP(c=k|\vv)\rp_{\btheta}(\vv|\vx) d\vv=\int \softmax(\vv)_k \rp_{\btheta}(\vv|\vx) d\vv,
\end{equation}
where $\rp_{\btheta}(\vv|\vx)$ is the marginal likelihood of the base CNF defined in Eqn.~(\ref{eq: cnf p(y|x)}).
Since the marginal likelihood of the S-CNF given in Eqn.~(\ref{eq: S-CNF-mlkl}) is analytically intractable due to the softmax transformation, we propose to approximate it using variational inference \citep{wainwright2008graphical} with a learnable mean-field Gaussian variational posterior $\rq_\bOmega(\vv|\vy)=\mathcal{N}(\vv|\bnu_\bOmega(\vy), \diag(\bsigma_\bOmega^2(\vy)))$, which can be seen as a probabilistic inverse of the softmax transformation $\rp(\vy|\vv)$. Applying Jensen's inequality to the log marginal likelihood of the S-CNF in Eqn.~(\ref{eq: S-CNF-mlkl}), a tractable evidence lower bound (ELBO) is obtained:
\begin{equation}\label{eq:s-cnf elbo}
    \log \rP_\btheta(c=k|\vx) 
    \geq \mathds{E}_{\rq_\bOmega(\vv|\vy)} \left[\log \rP(c=k|\vv) + \log \rp_\btheta(\vv|\vx) - \log \rq_\bOmega(\vv|\vy) \right ].
\end{equation}
It is worth noting that the softmax flow likelihood $\rP(c=k|\vv)=\softmax(\vv)_k$ places non-zero probability mass for every category $k=1,\cdots,K$, which is different from argmax flow~\citep{NEURIPS2021_67d96d45} whose likelihood only places probability mass for a single category. From a modelling perspective, softmax flow has a better capacity to represent the variability and uncertainty in human annotations. 
From an optimization perspective, the ELBO for softmax flow is always well-defined, whereas the ELBO for argmax flow is not defined when the model output does not match the human annotation, for which the reason lies in the fact that the log-likelihood would be $\log(0)$ in this case, which requires additional thresholding tricks to fix \citep{NEURIPS2021_67d96d45}.

\textbf{Meta-learning.} Using the variational approximation defined in Eqn.~(\ref{eq:s-cnf elbo}), the loss $\mathcal{L}(\btheta,\bOmega;\mathcal{D}_i,\vx_i)$ for S-CNF on a single dataset $\mathcal{D}_i$ can be defined as the average negative ELBO evaluated on the human annotations $\mathcal{D}_i=\{\bfeta_i^{(m)}\}_{m=1}^{M_i}$ given the corresponding input $\vx_i$:
\begin{equation}
    \small
    \mathcal{L}(\btheta,\bOmega;\mathcal{D}_i,\vx_i)  = -  \frac{1}{M_i}\sum_{m=1}^{M_i} \mathbb{E}_{\rq_\bOmega(\vv|\bfeta_{i}^{(m)} )}\left[\sum_{k=1}^K  \bfeta_{i,k}^{(m)} \log \rP(c_i=k|\vv) + \log \rp_\btheta(\vv|\vx_i) - \log \rq_\bOmega(\vv| \bfeta_{i}^{(m)}) \right],
\end{equation}
where the expectation over the variational posterior is approximated by Monte Carlo simulation with the reparameterization trick~\citep{kingma2013auto}. As in Section~\ref{ssec:i-cnf},  we follow the episodic training scheme with a meta-learning objective $\mathcal{L}(\btheta,\bOmega;\mathcal{D}_\text{meta},\{\vx_i\}_{i=1}^N)=\mathbb{E}_{D_i\sim\rp(\mathcal{D})}[\mathcal{L}(\btheta,\bOmega;\mathcal{D}_i,\vx_i)]$ for meta-training and use a similar flow sampling scheme but apply the softmax function $\vy_*^{(m)}=\softmax(\vv_*^{(m)})$ to the samples $\vv_*^{(m)}$ from the base CNF at test time. Note that each sample of S-CNF is a categorical distribution with probabilities $\vy_*^{(m)}$.

\section{Evaluation Metrics}\label{sec:metrics}
Several metrics are adopted to measure the empirical performance of HAS in terms of mean/majority prediction, distribution matching, and human variability simulation.

\textbf{Mean/majority prediction.} For ordinal annotations, the root mean squared error is used to evaluate the quality of the mean prediction for all test inputs: $\text{RMSE}^{\bar{y}}=\sqrt{\frac{1}{N}\sum_{i=1}^N(\Bar{y}_i-\Bar{\eta}_i)^2}$, where $\Bar{y}_i=\frac{1}{M_*}\sum_{m=1}^{M_*}y_i^{(m)}$ and $\Bar{\eta}_i=\frac{1}{M_i}\sum_{m=1}^{M_i}\eta_i^{(m)}$. For categorical annotations, the classification accuracy (Acc) for the majority vote is evaluated for all test inputs that have majority human annotations.

\textbf{Distribution matching.} The negative log likelihood (NLL) is used to evaluate how well the model estimates the human annotation distribution:
$\text{NLL}^\text{all}=-\frac{1}{N}\sum_{i=1}^N\left(\frac{1}{M_i}\sum_{m=1}^{M_i} \log \rp_\btheta(\bfeta_i^{(m)}|\vx_i)\right)$.

\textbf{Inter-annotator disagreement simulation.} Apart from evaluating the goodness of fit, additional metrics are adopted to explicitly measure how well the model simulates the variability and disagreements in human annotations: 1) the root mean squared error of the standard deviations of the annotations for all test inputs: $\text{RMSE}^s=\sqrt{\frac{1}{N}\sum_{i=1}^{N}\left(\sigma_i - s_i \right)^2}$, where $\sigma_i=\sqrt{\frac{1}{M_i}\sum_{m=1}^{M_i}(\eta_i^{(m)}-\bar{\eta}_i)^2}$ and $s_i=\sqrt{\frac{1}{M_*}\sum_{m=1}^{M_*}(y_i^{(m)}-\bar{y}_i)^2}$ for ordinal annotations, and $\sigma_i=\frac{1}{K}\sum_{k=1}^{K}\sqrt{\frac{1}{M_i}\sum_{m=1}^{M_i}(\eta_{i,k}^{(m)}-\bar{\eta}_{i,k})^2}$ and $s_i=\frac{1}{K}\sum_{k=1}^{K}\sqrt{\frac{1}{M_*}\sum_{m=1}^{M_*}(y_{i,k}^{(m)}-\bar{y}_{i,k})^2}$ for categorical annotations, and 2) the absolute error of the average standard deviations of the annotations for all test inputs: $\mathcal{E}(\Bar{s})=|\bar{\sigma}-\bar{s}|$, where $\bar{\sigma}=\sum_{i=1}^N \sigma_i$ and $\bar{s}=\sum_{i=1}^N s_i$. For categorical annotations, Fleiss's kappa ($\kappa$)~\citep{fleiss1971measuring} is additionally adopted to measure the inter-annotator disagreements, where $\kappa$ is a real number between $-1$ and $+1$, with $-1$ indicating no observed agreement and $+1$ indicating perfect agreement. The absolute error between the kappas of human annotations ($\kappa$) and simulated annotations ($\hat{\kappa}$) for all test inputs is reported: $\mathcal{E}(\hat{\kappa})=|\hat{\kappa}-\kappa|$.

\section{Experiments}
\label{sec: exp}

\textbf{Setup.} The proposed meta-learned zero-shot density estimation method for HAS from Section~\ref{sec: CNF} is evaluated by three representative real-world human evaluation tasks for speech and natural language processing: emotion category annotation, toxic speech detection, and speech quality assessment. The results are reported using evaluation metrics defined in Section~\ref{sec:metrics} and  several representative examples are visualized, which demonstrate the superior capability of the proposed method to capture the aggregated behaviours of human annotators, match the distribution of human annotations, and simulate the variability and disagreement of human perception and interpretation.

\textbf{Baselines.} The proposed I-CNF and S-CNF are compared to baselines of various types such as ensemble methods, Bayesian methods, and conditional generative models. This includes deep ensemble (Ensemble)~\citep{lakshminarayanan2017simple}, Monte-Carlo dropout (MCDP)~\citep{gal2016dropout},  Bayes-by-backprop (BBB)~\citep{blundell2015weight}, conditional variational autoencoder (CVAE)~\citep{kingma2013auto}, conditional argmax flow (A-CNF)~\citep{NEURIPS2021_67d96d45}, Dirichlet prior network (DPN)~\citep{malinin2018predictive}, Gaussian process (GP)~\citep{williams2006gaussian}, and evidential deep learning (EDL)~\citep{amini2020deep}. 
We fit them to all available human annotations for all utterances in the training set, tune hyperparameters on the validation set, and report performance on the test set. $M_*=100$ samples are used to compute  evaluation metrics at test time. The Ensemble only consists of 10 systems due to its expensive computational cost.

\textbf{Backbone architecture.} The same neural network feature encoder is used in all compared methods to extract features from inputs, which follows an upstream-downstream paradigm. The upstream model, also called a foundation model~\citep{bommasani2021opportunities}, is pre-trained on a large amount of unlabelled data to learn universal representations. WavLM~\citep{chen2022wavlm} and RoBERTa~\citep{liu2019roberta} are used as the pre-trained upstream models for speech and text inputs, respectively. The downstream model consists of two Transformer encoder blocks followed by two fully connected (FC) layers, which are fine-tuned to target specific applications.

\begin{table}[b]
\footnotesize
\centering
\caption{Test performance on the emotion category annotation task. CVAE collapses to one category for all inputs.}
\label{tab: emo-baseline}
\begin{tabular}{cccccc}
 \toprule
             & Acc$~(\uparrow)$   & $\text{NLL}^\text{all}~(\downarrow)$ & 
             $\text{RMSE}^s~(\downarrow)$ &
             $\mathcal{E}(\Bar{s})~(\downarrow)$ & $\mathcal{E}(\hat{\kappa})~(\downarrow)$\\
\midrule
MCDP     & 0.582$\pm$0.003&	\underline{1.423$\pm$0.012}&	0.294$\pm$0.001&	0.193$\pm$0.000&	0.467$\pm$0.005\\
Ensemble & \textbf{0.603$\pm$0.002}&	1.458$\pm$0.004&	0.271$\pm$0.003&	0.160$\pm$0.004&	0.344$\pm$0.017\\
BBB	& 0.565$\pm$0.010&	1.459$\pm$0.011&	0.289$\pm$0.005	&0.187$\pm$0.008&	0.511$\pm$0.034\\
DPN & 0.581$\pm$0.006&	1.518$\pm$0.002&	0.296$\pm$0.001&	0.193$\pm$0.001&	\underline{0.104$\pm$0.016}\\
CVAE & 0.275$\pm$0.000& 1.661$\pm$0.000&0.333$\pm$0.000& 0.244$\pm$0.000&	--- \\
A-CNF &      0.583$\pm$0.002&	1.430$\pm$0.006&	\underline{0.239$\pm$0.001}&	\underline{0.097$\pm$0.002}&	0.382$\pm$0.015\\
S-CNF         & \underline{0.591$\pm$0.002}&	\textbf{1.403$\pm$0.011}&	\textbf{0.218$\pm$0.000}&	\textbf{0.020$\pm$0.002}&
\textbf{0.068$\pm$0.021}\\
\bottomrule
\end{tabular}
\end{table}

\begin{figure}[tb]
    \centering
    \includegraphics[width=0.45\textwidth]{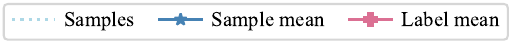}
\begin{minipage}[b]{\linewidth}
\centerline{\includegraphics[width=\linewidth]{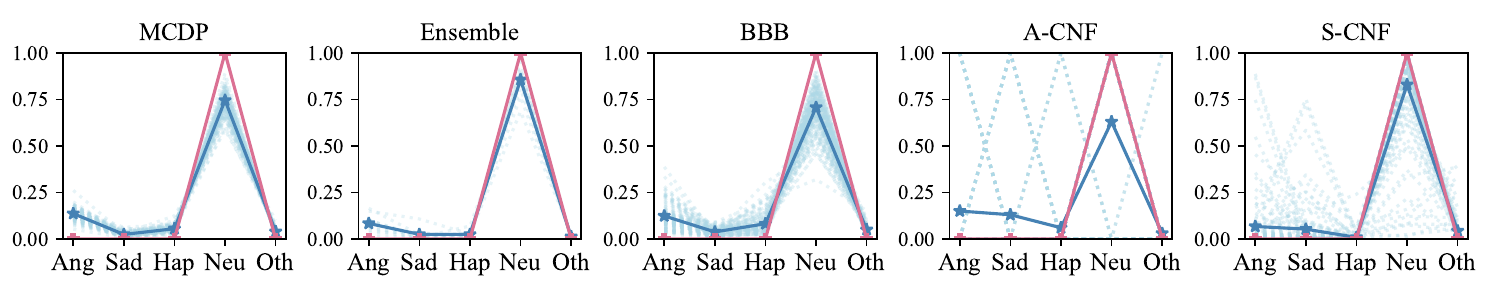}}
\small \centerline{(a) ``MSP-PODCAST\_0114\_0263.wav''}
    \end{minipage}
    \begin{minipage}[b]{\linewidth}
\small \centerline{\includegraphics[width=\linewidth]{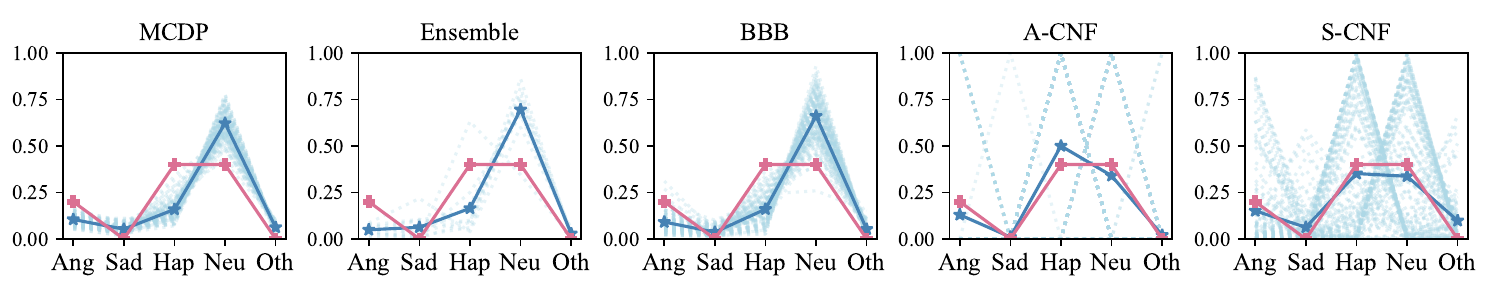}}
\centerline{(b) ``MSP-PODCAST\_0167\_0179\_0001.wav''}
    \end{minipage}
    \begin{minipage}[b]{\linewidth}
\centerline{\includegraphics[width=\linewidth]{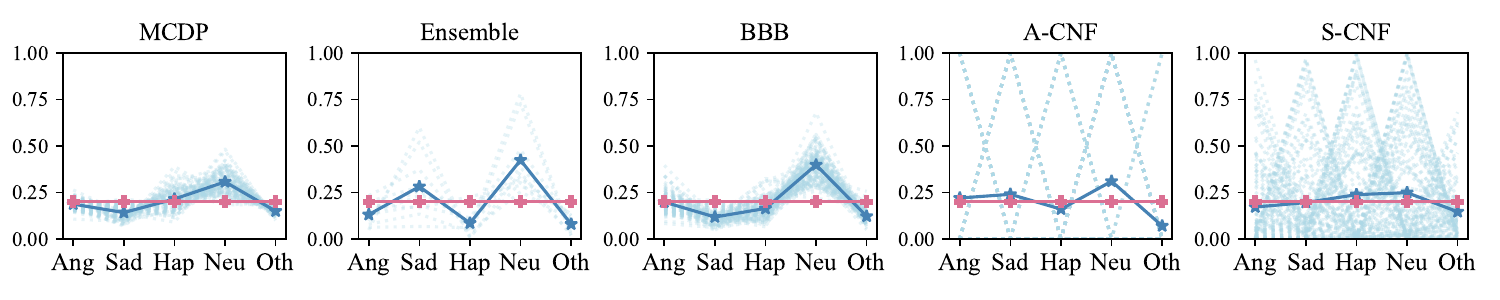}}
\small \centerline{(c) ``MSP-PODCAST\_0574\_0476.wav''}
    \end{minipage}
    \caption{Visualization of simulated annotations on the emotion category annotation task for case study. The y-axis corresponds to the probability mass. Each sample is a categorical distribution.
    The probability mass values of different categories in each categorical distribution are connected for the purpose of better visualization. CVAE is omitted because it collapses to one category for all inputs.}
    \label{fig: emo visual}
     
\end{figure}

\subsection{Emotion Category Annotation}
\label{sec: exp emo}

\textbf{Task.} Emotion recognition aims to identify human emotion, which is beneficial for healthcare, education and customization purposes. Human emotion is inherently ambiguous and the perception of emotion is highly subjective, which often results in disagreements among human annotators.
Most emotion recognition datasets employ multiple annotators to label each utterance. However, prior work typically uses the majority vote as the ground-truth target~\citep{Busso2008IEMOCAPIE, MSP-podcast,Meld}, which ignores minority viewpoints and thus fails to represent the true human annotation distributions.
Our proposed method can enhance the fairness of emotion category annotation as it better handles different opinions among human annotators.

\textbf{Dataset.} MSP-Podcast~\citep{MSP-podcast} is one of the largest publicly available datasets in speech emotion recognition, which contains natural English speech from podcast recordings annotated using crowd-sourcing. The experiment uses Release 1.6 of this dataset, which contains more than 50,000 utterances from more than 1,000 speakers consisting of more than 80 hours of speech. The standard splits of training (34,280 segments), validation (5,958 segments) and test (10,124 segments) are used. Emotion labels are grouped into five categories: angry, sad, happy, neutral, and other. Each utterance is labelled by at least 5 human annotators, and there are 6.7 annotations per utterance on average. It is worth noting that 16.5\% of the utterances in this dataset do not have a majority emotion class, showing strong disagreement among human annotators.

\textbf{Performance.} Table~\ref{tab: emo-baseline} reports the test results for all compared methods. Ensemble achieves the best majority prediction accuracy (Acc) at the cost of training 10 independent systems. The proposed S-CNF achieves the second-best majority prediction accuracy with only a tenth of the computational cost of Ensemble during training. More importantly, S-CNF is the best at matching the distributions of human annotations (in terms of $\text{NLL}^\text{all}$) and simulating inter-annotator disagreements (measured by $\text{RMSE}^s$, $\mathcal{E}(\Bar{s})$ and $\mathcal{E}(\hat{\kappa})$) among all compared methods.

\textbf{Case study.} To better illustrate the properties of the annotations simulated by different methods, The simulated distributions against the ground-truth distributions for three representative examples are visualized in Figure~\ref{fig: emo visual} (more case study examples can be found in Appendix~\ref{apdx: cat graph}). Overall, the mean of the samples generated by S-CNF aligns the best with the average human label, indicating its superior performance in estimating the aggregated behaviours of human annotators. Interestingly, the samples generated by S-CNF are the most diverse among all compared methods, which manage to simulate the variability of the behaviours of different individual human annotators. In sharp contrast, the samples generated by all the other methods highly concentrate around their sample means. The visualized result for each example is analyzed below:
\begin{enumerate}[label=(\alph*),topsep=0pt,itemsep=0pt]
    \item Human annotators reach a consensus in this case. The majority of samples generated by S-CNF exhibit prominent peaks aligned with the ground-truth emotion class ``neutral''. In contrast, many samples generated by A-CNF peak at other emotion classes.
    \item Human opinions diverge in this case. The majority of samples generated by S-CNF are sharp categorical distributions peaking at one of the two majority emotion classes ``happy'' and ``neutral''. Additionally, a few samples generated by S-CNF peak at the emotion class ``angry'', which manages to simulate the minority viewpoint held by some annotators. Very few human annotators attribute this utterance to the emotion classes ``sad'' and ``other'', and S-CNF likewise produces scarce samples peaking at these classes.
    \item Five human annotators give distinct emotion labels in this case, resulting in a tie in the label means. The tie comes from annotators' diverse individual perceptions of the emotion rather than consensus on its ambiguity. S-CNF is the only model that can simulate both the diverse behaviours of different individual annotators and the aggregated behaviour of all annotators since the individual samples are sharp categorical distributions peaking at one of the five emotion classes and the mean of the samples aligns well with the label mean.
\end{enumerate}

\subsection{Toxic Speech Detection}

\textbf{Task.} Toxic speech detection aims to filter out harmful and offensive language in written or spoken communications, such as insults, threats and harassment, which can lead to emotional distress, cyberbullying, and hostile online environments. 
Developing effective toxic detection methods is crucial for creating safer and more respectful online environments and promoting positive interactions and healthy communications among users.
Our proposed method incorporates interpretations from different human annotators, leading to a comprehensive understanding of hate speech, which is a good substitute for human annotators to reduce their exposure to distressing and harmful content.

\textbf{Dataset.} The HateXplain dataset~\citep{mathew2021hatexplain} is used in this experiment, which contains over 20,000 text posts from Twitter and Gab. 
These posts are labelled using crowd-sourcing with the commonly used 3-category annotation: hate, offensive, normal. Each post is annotated by three annotators. Cases where all the three annotators choose a different class (919 out of 20,148 posts) were originally excluded from the standard split of the dataset. We incorporate these cases into our training, validation, and test sets in an 8:1:1 ratio to better reflect the inter-annotator disagreements, resulting in 16,118 posts for training, 2,014 for validation, and 2,016 for testing in total. 

\textbf{Performance.} Table~\ref{tab: toxic-speech-baseline} reports the test results for all compared methods, which shows a similar trend to that found in emotion category annotation experiment in Section~\ref{sec: exp emo}. The Ensemble achieves the best majority prediction accuracy at the cost of training 10 independent systems. Our proposed S-CNF achieves the second best majority prediction accuracy while being much more computationally efficient and has the best performance in distribution matching and inter-annotator disagreement simulation. A case study with visualization can be found in Appendix~\ref{apdx: hate graph}, which also exhibits similar trends to those for the emotion category annotation experiment in Section \ref{sec: exp emo}.

\begin{table}[b]

\footnotesize
\centering
\caption{Test performance on the toxic speech detection task. CVAE collapses to one category for all inputs.}
\label{tab: toxic-speech-baseline}
\begin{tabular}{cccccc}
 \toprule
  & Acc$~(\uparrow)$   & $\text{NLL}^\text{all}~(\downarrow)$ & 
             $\text{RMSE}^s~(\downarrow)$ &
             $\mathcal{E}(\Bar{s})~(\downarrow)$ & $\mathcal{E}(\hat{\kappa})~(\downarrow)$\\
\midrule
MCDP     & 0.656$\pm$0.009 &	0.951$\pm$0.032&	0.300$\pm$0.002&	0.129$\pm$0.003	&0.143$\pm$0.008\\
Ensemble & \textbf{0.682$\pm$0.002}&	0.909$\pm$0.012&	\underline{0.289$\pm$0.001}&	0.100$\pm$0.003&	\underline{0.064$\pm$0.006}\\
BBB	& 0.670$\pm$0.001&	0.949$\pm$0.021&	0.300$\pm$0.009 &	0.127$\pm$0.022 &	0.207$\pm$0.051\\
DPN & 0.581$\pm$0.006&	1.158$\pm$0.002&	{0.296$\pm$0.001}&	0.193$\pm$0.001&	0.104$\pm$0.016\\
CVAE & 0.406$\pm$0.000 & 1.150$\pm$0.000 & 0.345$\pm$0.000 & 0.208$\pm$0.000 & --- \\
A-CNF &      0.628$\pm$0.003&	\underline{0.892$\pm$0.011}&	0.297$\pm$0.001&	\underline{0.087$\pm$0.008}&	0.198$\pm$0.027\\
S-CNF         & \underline{0.673$\pm$0.002}&	\textbf{0.837$\pm$0.008}&	\textbf{0.263$\pm$0.001}&	\textbf{0.002$\pm$0.001}&	\textbf{0.026$\pm$0.012} \\
\bottomrule
\end{tabular}
\end{table}

\subsection{Speech Quality Assessment}
\label{sec: MOS}

\textbf{Task.} Speech quality assessment plays an important role in the development of speech processing systems such as text-to-speech (TTS) synthesis. Speech quality is a complex, subjective psychoacoustic outcome of human perception. The mean opinion score (MOS) is a commonly used metric to evaluate the speech quality in TTS, which is obtained by having human listeners rate the perceived quality of the synthesized speech on a numerical scale typically ranging from 1 to 5, where a higher score indicates better-perceived speech quality, then averaging the scores across all listeners.
However, estimating only the MOS (\textit{i.e.}, the average score) and ignoring the individual scores fails to take into account the subjective nature of individual preferences, perceptions and biases.
Our proposed method is a cost-effective alternative to the time-consuming and expensive human assessment of speech quality which models the subjectivity that different human listeners may have.

\textbf{Dataset.} The SOMOS dataset~\citep{maniati22_interspeech} is used in this experiment, which consists of 20,000 synthetic utterances generated from 200 TTS systems and is annotated using crowd-sourcing. Each audio segment is evaluated by at least 17 unique annotators out of 987 participated human annotators, and there are 17.9 annotations per segment on average. The human annotators were asked to evaluate the naturalness of each audio sample on a 5-point Likert scale from 1 (very unnatural) to 5 (completely natural). The standard split provided by the dataset is used, which contains 141,100 training segments, 3,000 validation segments and 3,000 test segments. 

\textbf{Performance.} Table~\ref{tab: reg-baseline} reports the test results for all compared methods.
Again, our proposed I-CNF obtains the best performance for distribution match and inter-annotator disagreement simulation while achieving the second-best performance for MOS prediction. 
\begin{table}[tb]
\footnotesize
\centering
\caption{Test performance on the speech quality assessment task.}
\label{tab: reg-baseline}
\begin{tabular}{cccccc}
\toprule
     & $\text{RMSE}^{\bar{y}}~(\downarrow)$ & $\text{NLL}^\text{all}~(\downarrow)$ & $\text{RMSE}^s~(\downarrow)$ & $\mathcal{E}(\Bar{s})~(\downarrow)$ \\ 
     \midrule
GP   & \textbf{0.359$\pm$0.001}&	1.693$\pm$0.000&	{0.472$\pm$0.000} &	{0.412$\pm$0.000}\\
EDL	& 0.449$\pm$0.023&	\underline{1.636$\pm$0.001}&	\underline{0.375$\pm$0.022}&	\underline{0.356$\pm$0.025}\\
MCDP &\underline{0.390$\pm$0.013}&	1.787$\pm$0.008&	0.783$\pm$0.035&	0.742$\pm$0.031\\
Ensemble  & 0.410$\pm$0.008&	1.858$\pm$0.000&	0.740$\pm$0.007&	0.704$\pm$0.006\\
BBB	&0.613$\pm$0.011&	1.934$\pm$0.015&	0.944$\pm$0.017&	0.918$\pm$0.017\\
CVAE &0.419$\pm$0.013&	1.703$\pm$0.022&	0.598$\pm$0.033&	0.561$\pm$0.035\\
I-CNF &\underline{0.392$\pm$0.016}&	\textbf{1.609$\pm$0.003}&	\textbf{0.251$\pm$0.007}&	\textbf{0.123$\pm$0.013} \\ 
\bottomrule
\end{tabular}
\end{table}

\begin{figure}[b]
    \centering
    \includegraphics[width=0.31\linewidth]{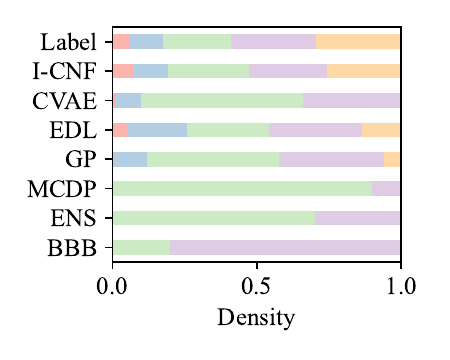}
    \includegraphics[width=0.31\linewidth]{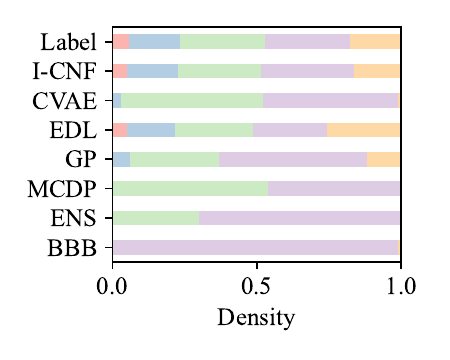}
    \includegraphics[width=0.31\linewidth]{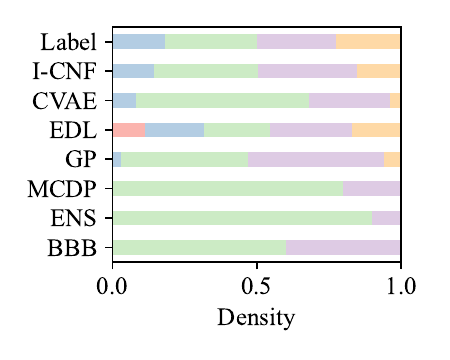}
    \includegraphics[width=0.04\linewidth]{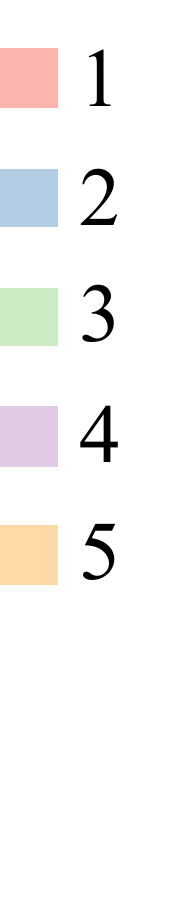}
    ~\\
    \begin{minipage}[b]{0.31\linewidth}
    \centerline{(a) ``booksent\_2013\_0010\_051''}
\end{minipage}
\begin{minipage}[b]{0.31\linewidth}
    \centerline{(b) ``news\_2010\_0183\_165''}
\end{minipage}
\begin{minipage}[b]{0.34\linewidth}
    \centerline{(c) ``reportorial\_2011\_0245\_098''}
\end{minipage}
    \caption{Visualization of simulated annotations on the speech quality assessment task for case study. The length of the bar in each colour represents the density of the corresponding score.}
    \label{fig: case study mos}    
\end{figure}

\textbf{Case study.} To better illustrate the properties of the annotations simulated by different methods, simulated distributions are visualized against the ground-truth distributions for three representative examples in Figure~\ref{fig: case study mos}. It can be seen that the proposed I-CNF is the only method which gives an accurate distribution match and perfect inter-annotator disagreement simulation in all three cases. In contrast, all the other methods tend to either produce annotations centered around the mean score or collapse to one score (typically 3 or 4). More case study examples can be found in Appendix~\ref{apdx: reg graph}.

\section{Conclusion}
This paper studied human annotator simulation (HAS), a cost-effective alternative to generating human-like annotation for automatic data labelling and model evaluation. 
To incorporate the variability of human evaluations, a novel framework was introduced which treats HAS as a zero-shot density estimation problem. This overcame the drawbacks of prior work and enabled efficient annotation simulation for unlabelled test inputs.
Under this framework, a meta-learning objective was derived for two new model classes, conditional integer flows and conditional softmax flows, to account for ordinal and categorical annotations, respectively.
The proposed method consistently and significantly outperformed a wide range of methods on three real-world human evaluation tasks, achieving the best performance for human annotation distribution matching and inter-annotator disagreement simulation. 
It is hoped that our proposed method could help mitigate unfair biases and over-representation in HAS and reduce the exposure of human annotators to potentially harmful content, thus promoting ethical AI practices.

\section*{Acknowledgements}
W. W. is supported by a Cambridge International Scholarship from the Cambridge Trust. W. C. acknowledges funding via a Cambridge Trust Scholarship and a Cambridge University Engineering Department Studentship under grant G105682 NMZR/089 supported by Huawei R\&D UK. This work has been performed using resources provided by the Cambridge Tier-2 system operated by the University of Cambridge Research Computing Service (www.hpc.cam.ac.uk) funded by EPSRC Tier-2 capital grant EP/T022159/1.

The MSP-Podcast data was provided by The University of Texas at Dallas through the Multimodal Signal Processing Lab. This material is based upon work supported by the National Science Foundation under Grants No. IIS-1453781 and CNS-1823166. Any opinions, findings, and conclusions or recommendations expressed in this material are those of the author(s) and do not necessarily reflect the views of the National Science Foundation or The University of Texas at Dallas.

\section*{Ethics Statement}
In this work, all human annotations used for training were taken from existing publicly available corpora, and no new human annotations were collected. 

It is hoped that this work could play a part in promoting ethical AI practice.
Firstly, 
it has been shown that the proposed HAS system can capture the inherent variability in human judgements and help mitigate biases and the issue of over-representation, thus producing a more inclusive representation of human opinions.
The proposed HAS system also has the potential to minimize human annotators' exposure to offensive and/or hateful content in some evaluation tasks such as HateXplain.

 \section*{Reproducibility Statement}
The datasets used in the experiments are all publicly available. The source code associated with the proposed method can be found at \url{https://github.com/W-Wu/HAS_CNF}.

\bibliography{iclr2024_conference}
\bibliographystyle{iclr2024_conference}

\clearpage
\section*{{\Large Appendices}}
\appendix
\section{The Sources of Variability in Human Evaluation}
\label{apdx: variability in HAS}

Human perception refers to the process by which individuals interpret and make sense of the sensory information they receive from the environment. It involves the integration of sensory data, cognitive processes, emotions, and previous experiences. Subjective perception emphasizes that each individual's perception of the world is unique and influenced by their internal mental states, beliefs, attitudes, and past experiences. As a result, people can interpret and react to the same stimuli differently, leading to diverse and subjective perceptions.

Each person's sensory organs, such as eyes and ears, may have slight variations in sensitivity and acuity, leading to different perceptions of the same stimuli. Cognitive biases, the inherent mental shortcuts or tendencies that influence how humans perceive and process information, can lead to difference in judgement and decision-making. People's past experiences, cultural norms, and upbringing also shape their perceptions. Different cultural backgrounds can lead to distinct interpretations of the same event, leading to diverse reaction. The variability in humans can manifest in various tasks such as colour perception, emotion recognition, art appreciation, and feedback preferences.

Embracing and understanding the variability of human perception is vital for various research fields such as psychology, neuroscience, human-computer interaction, and so on, and has practical implications in designing human-centered systems and promoting empathy and diversity. It helps create products and interfaces that cater to diverse user needs and preferences in fields like human-computer interaction and user experience design. Being aware of the variability of perception is crucial in ethical decision-making. It help ensures that different perspectives and cultural sensitivities are considered, which helps identify and address potential biases that might disproportionately affect certain groups or lead to unfair outcomes.

\section{Derivations}
\label{apdx: derivation}
Detailed derivations for the training objectives on a single dataset $\mathcal{D}_i=\{ \bfeta_i^{(1)}, \cdots ,  \bfeta_i^{(M)}\}$ with $\vx_i$ are presented in this section. For the simplicity of notations, the subscription $i$ in our derivations will be omitted without ambiguity where possible. The meta-learning objectives presented in the paper are obtained by averaging such single-task objectives across tasks.

\subsection{Objective Function for the Base CNF and I-CNF}
Denote the empirical human annotation distribution as $\rp_m(\vy|\vx)=\delta\left(\vy-\bfeta^{(m)}\right), \ m = 1, \cdots , M$ and model output distribution as $\rp_\btheta(\vy|\vx)$. The average KL divergence between them over all $M$ human annotations for this input $\vx$ is given by:
\begin{equation}
    \begin{aligned}
    \mathcal{L}(\btheta;\mathcal{D},\vx)&=\frac{1}{M}\sum_{m=1}^M\mathcal{KL}\left(\rp_m(\vy|\vx) \parallel \rp_\btheta(\vy|\vx)\right)\\
    &= \frac{1}{M}\sum_{m=1}^M \int \rp_m(\vy|\vx) \log \frac{\rp_m(\vy|\vx)}{\rp_\btheta(\vy|\vx)} d\vy\\
    &= - \frac{1}{M}\sum_{m=1}^M \int \rp_m(\vy|\vx) \log \rp_\btheta(\vy|\vx) d\vy + \text{const} \\
    &= - \frac{1}{M}\sum_{m=1}^M \log \rp_\btheta(\bfeta^{(m)}|\vx) + \text{const}
\end{aligned}
\end{equation}
Minimizing this KL objective is equivalent to maximizing the average log likelihood $\log \rp_\btheta(\bfeta^{(m)}|\vx)$ over all human annotations as presented in the paper. With numerical approximation, the training objective for I-CNF shares the same formula as that for the base CNF.

\subsection{Objective Function of S-CNF}
\label{apdx: derivation-cat}

For categorical annotations, each label $\bfeta^{(m)}$ represents the probabilities of all categories in the categorical human annotation distribution: $\bfeta^{(m)}=[ \bfeta_1^{(m)}, \cdots ,  \bfeta_K^{(m)}]$, where $\bfeta_k^{(m)}=\rP_m(c=k|\vx)$. Denote the model output distribution as $\rP_\btheta(c|\vx)$. The average KL divergence between them over all $M$ human annotations for this input $\vx$ is given by:
\begin{equation}
    \begin{aligned}
    \mathcal{L}^{\text{exact}}(\btheta;\mathcal{D},\vx)&=\frac{1}{M}\sum_{m=1}^M\mathcal{KL}\left(\rP_m(c|\vx) \parallel \rP_\btheta(c|\vx)\right)\\
    &= \frac{1}{M}\sum_{m=1}^M\sum_{k=1}^K \rP_m(c=k|\vx) \log \frac{\rP_m(c=k|\vx) }{\rP_\btheta(c=k|\vx)} \\
    &= - \frac{1}{M}\sum_{m=1}^M\sum_{k=1}^K \rP_m(c=k|\vx) \log \rP_\btheta(c=k|\vx) + \text{const} \\
    &=-\frac{1}{M}\sum_{m=1}^M\sum_{k=1}^K  \bfeta_k^{(m)} \log \rP_\btheta(c=k|\vx) + \text{const},
\end{aligned}
\end{equation}
where the marginal likelihood is lower bounded using variational inference:
\begin{equation}
    \begin{aligned}
\log \rP_\btheta(c=k|\vx) &= \log \int \rP(c=k|\vv)\rp_\btheta(\vv|\vx) d\vv \\
&= \log \int \vq_\bOmega(\vv|\bfeta ) \frac{\rP(c=k|\vv)\rp_\btheta(\vv|\vx)}{\vq_\bOmega(\vv|\bfeta )}d\vv \\
&\geq \int \vq_\bOmega(\vv|\bfeta ) \log \frac{\rP(c=k|\vv)\rp_\btheta(\vv|\vx)}{\vq_\bOmega(\vv|\bfeta )}d\vv \\
&= \mathds{E}_{ \vq_\bOmega(\vv|\bfeta )} \left[\log \rP(c=k|\vv) + \log \rp_\btheta(\vv|\vx) - \log \vq_\bOmega(\vv|\bfeta ) \right ].
\end{aligned}
\end{equation}

Therefore, the final negative ELBO objective is obtained by
\begin{equation}
    \begin{aligned}
     \mathcal{L}^{\text{exact}} &=- \frac{1}{M }\sum_{m=1}^{M } \sum_{k=1}^K  \bfeta_k^{(m)}
     \log \rP_\btheta(c=k|\vx)\\
     &\leq -  \frac{1}{M }\sum_{m=1}^{M } \sum_{k=1}^K  \bfeta_k^{(m)} \mathds{E}_{\vq_\bOmega(\vv| \bfeta^{(m)})} \left[\log \rP(c=k|\vv) + \log \rp_\btheta(\vv|\vx) - \log \vq_\bOmega(\vv| \bfeta^{(m)}) \right ]\\
     &= -  \frac{1}{M }\sum_{m=1}^{M } \mathds{E}_{\vq_\bOmega(\vv| \bfeta^{(m)})} \left[\sum_{k=1}^K  \bfeta_k^{(m)} \log \rP(c=k|\vv) + \log \rp_\btheta(\vv|\vx) - \log \vq_\bOmega(\vv| \bfeta^{(m)}) \right ]\\
     & = \mathcal{L} (\btheta,\bLambda;\mathcal{D},\vx),
\end{aligned}
\end{equation}
where
\begin{align}
    &\log \rP(c=k|\vv) = \operatorname{logsoftmax}(\vv)_k,\\
    & \log \rp_\btheta(\vv|\vx) =\rp_\bLambda\left(\vf^{-1}_\btheta(\vv)|\vx\right) \left|\operatorname{det}\left(\frac{\partial \vf^{-1}_\btheta(\vv)}{\partial \vv} \right) \right|,    \\
    & \log \vq_\bOmega(\vv| \bfeta^{(m)}) = \mathcal{N}(\vv | \bnu_\bOmega( \bfeta^{(m)}), \diag(\bsigma_\bOmega^2( \bfeta^{(m)}))).
\end{align}

\subsection{The Negative Log Likelihood ($\text{NLL}^\text{all}_i$) for Categorical Annotations}
The marginal likelihood of S-CNF is intractable, which can be approximated using Monte-Carlo simulation:
\begin{equation}
    \begin{aligned}
         \rP_\btheta(c=k|\vx) 
    &= \int \rP(c=k|\vv) \rp_\btheta(\vv|\vx) d\vv \\
    &= \mathds{E}_{\rp_\btheta(\vv|\vx)} \left[\rP(c=k|\vv)\right]  \\
    &\approx \frac{1}{Q} \sum_{j=1}^Q  \rP(c=k|\vv_j), \quad \{\vv_j\}_{j=1}^Q \sim_\text{iid} \rp_\btheta(\vv|\vx)  \\
    &= \frac{1}{Q} \sum_{j=1}^Q  \softmax(\vv_j)_k, \quad \{\vv_j\}_{j=1}^Q \sim_\text{iid} \rp_\btheta(\vv|\vx)  \\
    &= \bar{\vy}_k,
    \end{aligned}
\end{equation}
    where $\bar{\vy}= \frac{1}{Q} \sum_{j=1}^Q \softmax(\vv_j) = \frac{1}{Q} \sum_{j=1}^Q \vy_{j}$ which is the average of the simulated categorical distributions. Let $\bar{ \bfeta}= \frac{1}{M} \sum_{m=1}^M  \bfeta^{(m)}$ be the average label. 
    
    Then, the $\text{NLL}_i^{\text{all}}$ for a single input $\vx_i$ is given by
    \begin{equation}
        \begin{aligned}
    \text{NLL}^\text{all}_i 
    &= -\frac{1}{M} \sum_{m=1}^M \sum_{k=1}^K  \bfeta_{i,k}^{(m)} \log \rP_\btheta(c=k|\vx_i) \\
    &\approx -\frac{1}{M} \sum_{m=1}^M \sum_{k=1}^K  \bfeta_{i,k}^{(m)} \log \bar{\vy}_{i,k} \\
    &= -\sum_{k=1}^K \bar{ \bfeta}_{i,k} \log \bar{\vy}_{i,k},  \label{eqn: nll-all-cat}
\end{aligned}
    \end{equation}
which is the cross entropy between the averaged label and averaged sample. 

\section{Emotion Label Processing for MSP-Podcast}
\label{apdx: emo class}
In MSP-Podcast, each annotator can choose from ten emotion classes to label the primary emotion of an utterance: \textit{Angry, Sad, Happy, Surprise, Fear, Disgust, Contempt, Neutral, Other}. Although only one option is allowed, they can say \textit{other} and define their own emotion class which can be more than one. During label processing, the original \textit{other} class is split into sub-classes depending on the manual defined label and merged with the pre-defined labels. The grouping details are shown as follows: (i) \textit{Angry} includes  \textit{angry, disgust, contempt, annoyed};  (ii)     \textit{Sad} includes \textit{sad, frustrated, disappointed, depressed, concerned}; (iii) \textit{Happy} includes \textit{happy, excited, amused}; (iv) \textit{Neutral} includes \textit{neutral}; (v)  \textit{Other} includes all other emotion subclasses not listed above.

\section{Model Structure Details}
\label{sec: struc detail}
\begin{figure}[tb]
    \centering
    \includegraphics[width=0.65\textwidth]{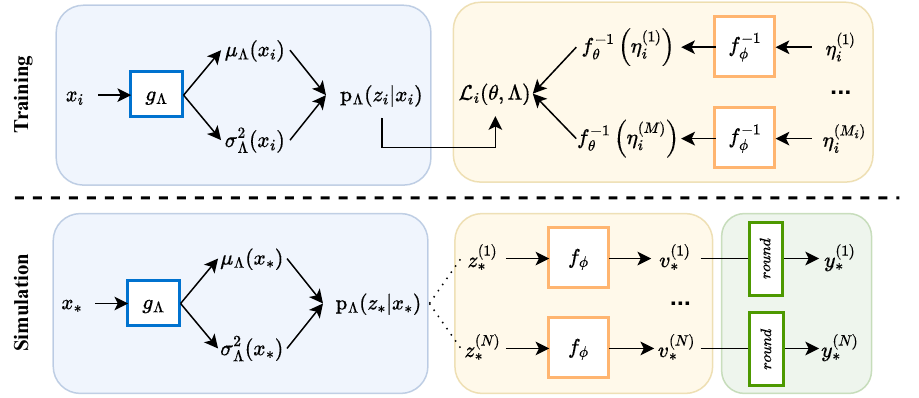}
    \caption{Illustration for I-CNF training and simulation workflow.}
    \label{fig: struc cnf}
\end{figure}

\begin{figure}[tb]
    \centering
    \includegraphics[width=\textwidth]{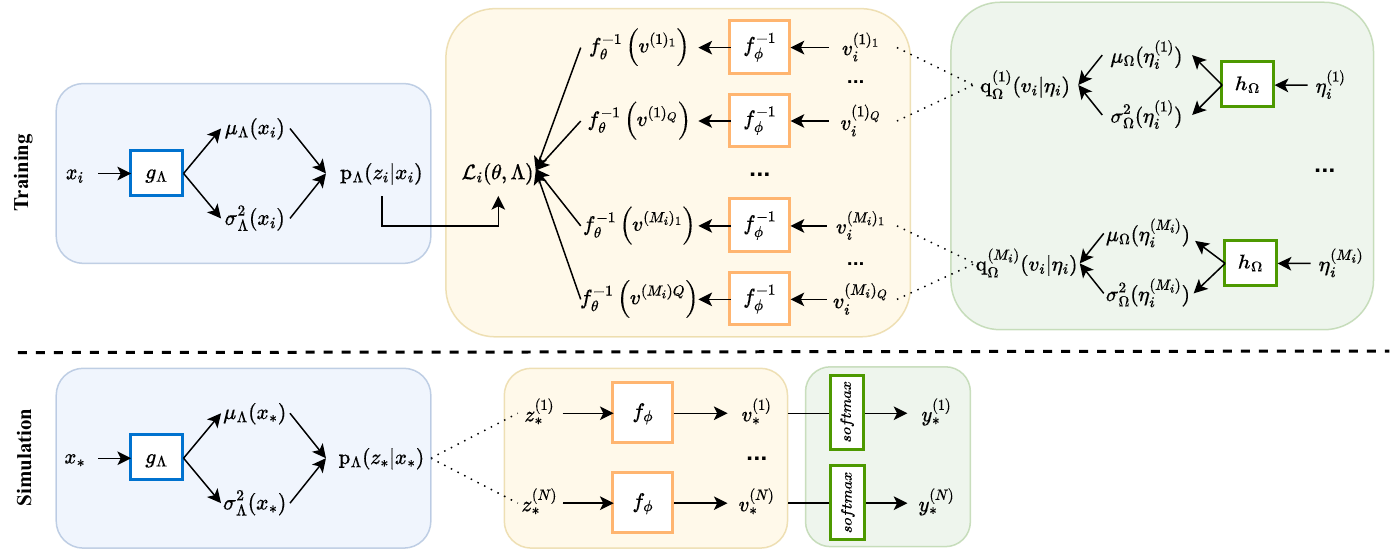}
    \caption{Illustration for S-CNF training and simulation workflow.}
    \label{fig: struc sf}
\end{figure}

The structure of proposed I-CNF and S-CNF are illustrated in Figure~\ref{fig: struc cnf} and Figure~\ref{fig: struc sf} respectively. The procedure of sampling from and optimizing S-CNF are summarized in Algorithm~\ref{al: sf-fwd} and~\ref{al: sf-bwd}.

\begin{minipage}[t]{0.46\linewidth}
\begin{algorithm}[H]
	\caption{Sampling from S-CNF} 
	\begin{algorithmic}
	    \State \textbf{Input}:  $\vx$
            \State \textbf{Output}: Categorical probability 
            $\vy$
	    \State  Compute $\bnu_\bLambda(\vx), \bsigma_\bLambda^2(\vx) = \vg_\bLambda(\vx)$
         \State Sample $\vz \sim \mathcal{N}(\bnu_\bLambda(\vx), \diag(\bsigma_\bLambda^2(\vx))$
     \State Compute $\vv = \vf_\btheta\left(\vz\right)$
     \State Compute $\vy = \softmax(\vv)$
	\end{algorithmic}
	\label{al: sf-fwd}
\end{algorithm}
\end{minipage}
\begin{minipage}[t]{0.54\linewidth}
\begin{algorithm}[H]
	\caption{Optimizing S-CNF} 
	\begin{algorithmic}
	    \State \textbf{Input}:  $\vx,  \mathcal{D}=\{\bfeta^{(1)}, \cdots ,  \bfeta^{(M)}\}$
            \State \textbf{Output}: ELBO 
            $\mathcal{L}^{\text{ELBO}}$ on dataset $\mathcal{D}$
            \For {$m=1, \cdots , M$}
    	    \State  Compute $\bnu_\bOmega( \bfeta^{(m)}), \bsigma_\bOmega^2( \bfeta^{(m)}) = \vh_\bOmega( \bfeta^{(m)})$
         \For {$j = 1, \cdots , Q$}
             \State Sample $\vv_j \sim \vq_\bOmega(\vv| \bfeta^{(m)})$
             \State Compute $\mathcal{L}_j^{(m)} = -\sum_{k=1}^K  \bfeta_k^{(m)} \log \rP(c=k|\vv_j) + \log \rp_\btheta(\vv_j|\vx) - \log \vq_\bOmega(\vv_j| \bfeta^{(m)})$
             \EndFor
             \State Compute $\mathcal{L}_m^{\text{ELBO}} =  \frac{1}{Q}\sum_{j=1}^Q\mathcal{L}_j^{(m)}$
         \EndFor
             
         \State Compute $\mathcal{L}^{\text{ELBO}} = \frac{1}{M} \sum_{m=1}^M \mathcal{L}_m^{\text{ELBO}}$
	\end{algorithmic}
	\label{al: sf-bwd}
\end{algorithm}
\end{minipage}

A neural-network-based encoder $\vg_\bLambda$ is built to model $\bnu_\bLambda(\vx), \bsigma_\bLambda^2(\vx)$ given input $\vx$ where $\bLambda$ is the model parameters. $\vg_\bLambda$ follows an upstream-downstream paradigm. The upstream model is pretrained on large amount of unlabelled data to learn universal representations. The downstream model uses the learned representation from the upstream model for specific applications. In this paper, the downstream model consists of two Transformer encoder blocks followed by two FC layers. The output layer contains two heads to predict the mean and standard deviation of the latent distribution $\rp_\bLambda(\vz|\vx)$.  

For tasks involving speech as input (\textit{i.e.}, emotion class labelling, speech quality prediction), WavLM~\citep{chen2022wavlm} is used as the upstream model. WavLM is a speech foundation model pre-trained by self-supervised learning that takes raw waveform as input. The waveform is encoded by a CNN encoder followed by multiple Transformer encoders. The BASE+ version\footnote{https://huggingface.co/microsoft/wavlm-base-plus} of the model is used in this work which has 12 Transformer encoder blocks with 768-dimensional hidden states and 8 attention heads. The parameters of the pretrained WavLM are frozen and the weighted sum of the outputs of the 12 Transformer encoder blocks is used as the speech embeddings feeding into the downstream model. 

RoBERTa~\citep{liu2019roberta} is used as upstream model to encode text input for toxic speech detection, which is a robustly optimized model of BERT~\citep{devlin-etal-2019-bert}. RoBERTa is a Transformer-based language model pretrained on a large corpus of English data with the masked language modelling objective. The BASE version\footnote{{https://huggingface.co/roberta-base}} was used in the work which has 12 Transformer layers, 768 hidden units, 12 attention heads, and 125 million parameters.

The downstream model consists of two Transformer encoder layers with hidden dimension of 128 and four attention head. The invertible flow model $\vf_\btheta$ uses real NVP block~\cite{dinh2016density}. The variational encoder for S-CNF $\vh_\bOmega$ contains a FC layer and two output heads for the mean and standard deviation of the variational distribution $\vq_\bOmega(\vv|\vy)$. More details can be found in Table~\ref{tab: struc detail}.

\begin{table}[tb]
\footnotesize
    \centering
    \caption{Dimension of the model structure (number of layers * layer dimension)}
    \label{tab: struc detail}
    \begin{tabular}{c|ccccc}
    \toprule
    Task     & input modality & $\vg_\bLambda$-upstream & $\vg_\bLambda$-downstream &  $\vf_\btheta$ & $\vh_\bOmega$\\
    \midrule
    Emotion class labelling & speech    & WavLM base+ & 2*128 & 3*64 &1*64\\
    Toxic speech detection & text  &   RoBERTa base   & 2*128 & 3*64 &1*64\\
    Speech quality assessment & speech & WavLM base+ & 2*128 & 3*16 & / \\
    \bottomrule
    \end{tabular}
\end{table}

\section{Detailed Configuration of All Compared Methods}
Ensemble consists of 10 systems initialized and trained using different random seeds. MCDP uses dropout rate of 0.4. A standard Gaussian prior is used for BBB. A modified version of EDL is used~\citep{wu-etal-2023-estimating} which is trained by maximising the per-observation-based marginal likelihood with a modified regularization term. Ensemble, MCDP, BBB, EDL use the same model structure as $\vg_\bLambda$ apart from removing the output head for predicting variance of latent distribution. 
A modified version of DPN\citep{wu2022estimating} is used which is trained by interpolating per-observation-based marginal likelihood with KL divergence. The coefficient of KL term is set to 5.0 for emotion class labelling and 2.0 for toxic speech detection. Features extracted from the upstream model are used as input to GP which uses a radial basis function kernel and is trained by maximising the per-observation-based marginal likelihood.  CVAE has the same $\vg_\bLambda$ structure as S-CNF for modelling $\rp(\vz|\vx)$, and two 64-d FC layers are used for encoder and decoder. A-CNF has identical model structure as S-CNF.

The system was implemented using PyTorch with the SpeechBrain~\citep{speechbrain} and normflows~\citep{Stimper2023} toolkit. The Adadelta optimizer was used with an initial learning rate of 1.2 for emotion class labelling and 0.05 for speech quality assessment. The The NewBob learning rate scheduler was used with annealing factor 0.8 and patience 1. The system was trained for 30 epochs and the model with the best validation performance was used for testing. The number of ELBO samples was set to 20.

\section{Analysis of Standard Deviation of Simulated Samples}
\label{apdx: std}
\begin{figure}[bt]
    \centering
        \includegraphics[width=0.9\linewidth]{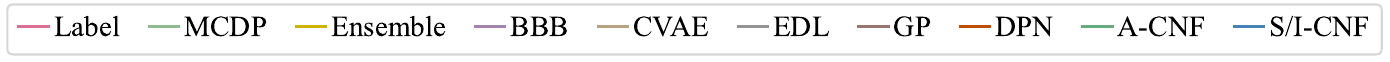}
    \begin{minipage}[b]{0.49\linewidth}
\centerline{\includegraphics[width=\linewidth]{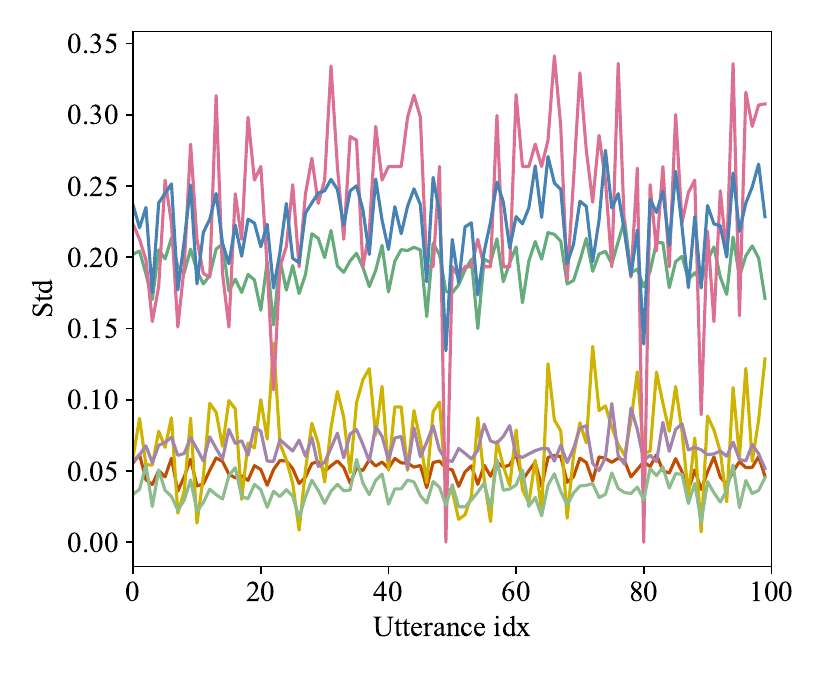}}
\centerline{(a) Std for emotion class labelling.}
    \end{minipage}
    \begin{minipage}[b]{0.49\linewidth}
\centerline{\includegraphics[width=\linewidth]{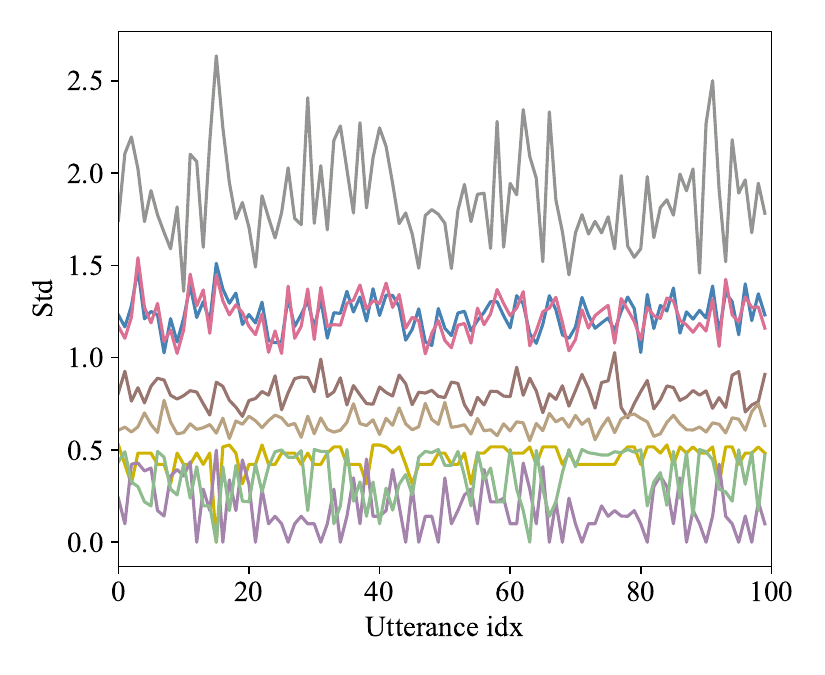}}
\centerline{(b) Std  for speech quality evaluation. }
    \end{minipage}
    \caption{Standard deviation of simulated samples.}
    \label{fig: std}
\end{figure}

It has been observed in Section~\ref{sec: exp emo} that flow models tend to have a larger difference between $\text{RMSE}^s$ and $\mathcal{E}(\Bar{s})$. This section provides detailed analysis to this observation. Let
$N$ be the number of test utterances. Three std-related metrics are computed: (i) RMSE between std of predictions and human labels: $\text{RMSE}^s=\sqrt{\frac{1}{N}\sum_{i=1}^{N}\left(s_i - \sigma_i \right)^2}$; (ii) Mean absolute error between std of predictions and std of human labels: $\text{MAE}^s=\frac{1}{N}\sum_{i=1}^N| s_i -\sigma_i | $; (iii)  Absolute error between average std of predictions and average std of human labels $\mathcal{E}(\Bar{s}) = \left|  \bar{s}_i -\bar{\sigma}_i\right|$. Results are shown in Table~\ref{tab: std analysis}.
\begin{table}[tb]
\footnotesize
\centering
\caption{Analysis of standard deviation of simulated samples}
\label{tab: std analysis}
\begin{tabular}{c|ccc|ccc|ccc}
\toprule
& \multicolumn{3}{c|}{Emotion recognition} & \multicolumn{3}{c|}{Toxic detection} & \multicolumn{3}{c}{Speech quality} \\
             & $\text{RMSE}^s$ &
             $\text{MAE}^s$ & 
             $\mathcal{E}(\Bar{s})$ & 
             $\text{RMSE}^s$ &
             $\text{MAE}^s$ & 
             $\mathcal{E}(\Bar{s})$ & $\text{RMSE}^s$ &
             $\text{MAE}^s$ & 
             $\mathcal{E}(\Bar{s})$ \\
             \midrule
MCDP     & 0.305&0.233&0.206   & 0.297 & 0.242 & 0.122 &0.809&0.762&0.762
       \\
Ensemble & 0.277&0.222&0.166 & 0.290&0.220&0.105 & 0.747&0.703&0.703 \\
BBB & 0.284&0.226&0.178& 0.279&0.229&0.115 & 0.952&0.917&0.917 \\
CVAE & 0.333&0.244&0.244 & 0.345&0.208 &0.208 & 0.574&0.535&0.534 \\
EDL & \multicolumn{3}{c|}{/}&\multicolumn{3}{c|}{/} & 0.381&0.368&0.368 \\
GP & \multicolumn{3}{c|}{/} & \multicolumn{3}{c|}{/} & 0.472&0.419&0.412 \\
DPN & 0.297&0.236&0.191 & 0.299&0.220&0.178 & \multicolumn{3}{c}{/}\\
A-CNF & 0.223&0.209&0.046 & 0.274&0.232&0.062 & \multicolumn{3}{c}{/} \\
S/I-CNF         & 0.218     & 0.198         & 0.015  & 0.263 & 0.206 & 0.002 & 0.229&0.184&0.067
       \\
\bottomrule
\end{tabular}
\end{table}
The flow model tends to have larger discrepancy between  $\text{MAE}^s$ and $\mathcal{E}(\Bar{s})$. According to the triangular inequality:
\begin{equation}
    \begin{aligned}
    \mathcal{E}(\Bar{s})=\left| \frac{1}{N}\sum_{i=1}^N s_i - \frac{1}{N}\sum_{i=1}^N \sigma_i \right|
    =\left| \frac{1}{N}\sum_{i=1}^N (s_i - \sigma_i) \right| 
    \leq \frac{1}{N}\sum_{i=1}^N\left|  s_i - \sigma_i\right| = \text{MAE}^s
\end{aligned}
\end{equation}
which show that $\mathcal{E}(\Bar{s})$ is a lower bound of $\text{MAE}^s$. The equality condition is satisfied when all samples are uniformly either greater than or less than the compared value. Therefore, a larger discrepancy between these two values indicates that the standard deviation of some samples exceeds that of the labels, while for others, it is lower. A smaller discrepancy indicates that the standard deviation of samples tend to be consistently larger of smaller than that of the labels. In Figure~\ref{fig: std}, 100 test utterances are randomly selected and the std of samples generated by different models are plotted, which supports the above conclusion. The proposed S-CNF and I-CNF has the best performance for matching the diversity of human annotations.

\section{Adjusting Diversity of CNFs by Prior Tempering}
\label{apdx: diversity}
\begin{figure}[tb]
    \centering
    \includegraphics[width=\textwidth]{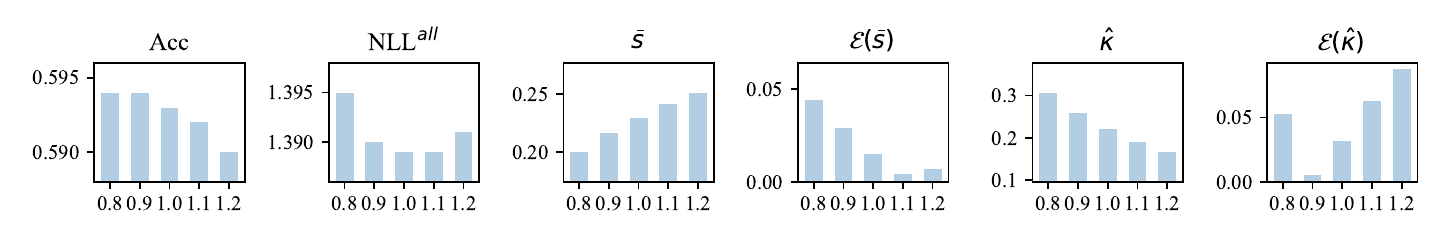}
    \centerline{(a) Emotion class labelling}
    \includegraphics[width=\textwidth]{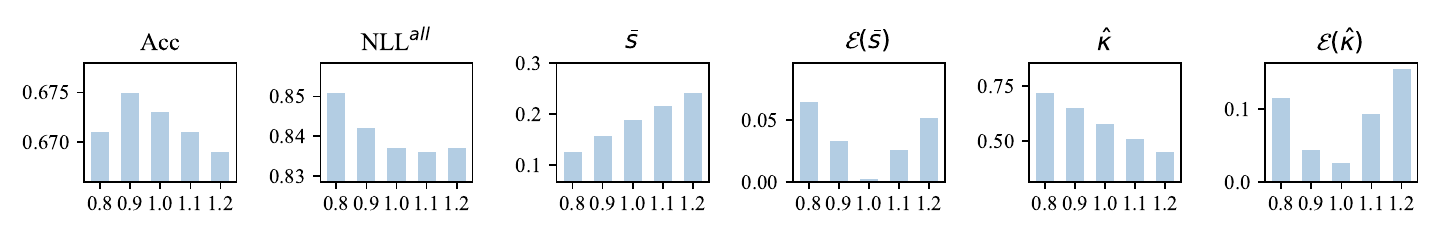}
    \centerline{(b) Toxic speech detection}
    \flushleft \includegraphics[width=0.67\textwidth]{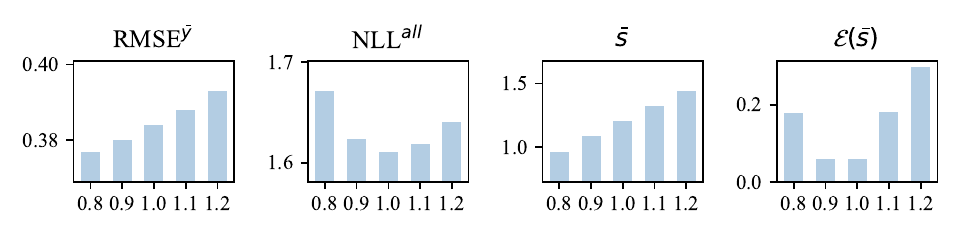}
    \centerline{(c) Speech quality assessment}
    \caption{The effect of prior tempering on the performance of S-CNF and I-CNF. The x-axis corresponds to the prior temperature.}
    \label{fig: T-softmax-appendix}
\end{figure}

One advantage of CNF is that its sample diversity can be easily controlled on demand without re-training by tempering the standard deviation of $\rp_\bLambda(\vz|\vx)$ at test time.
Figure~\ref{fig: T-softmax-appendix} explores the effect of prior tempering on the performance. More details are shown in Table~\ref{tab: flowT}. Overall, the trend is clear that the simulated annotations become more diverse as the temperature increases. 
The default temperature value $1$ used during training (\textit{i.e.}, no tempering) achieves the best trade-off among majority prediction accuracy (Acc), distribution matching ($\text{NLL}^\text{all}$), and inter-annotator disagreement simulation (in terms of $\mathcal{E}(\Bar{s})$ and $\mathcal{E}(\hat{\kappa})$). In addition, as compared in Table~\ref{tab: mcdp diversity}, prior tempering in CNF is more efficient and covers a wider range of dynamics than adjusting the dropout rate in MCDP.

\begin{table}[tb]
\footnotesize
\centering
\caption{Adjusting the diversity of CNFs by prior tempering.}
\label{tab: flowT}
\begin{tabular}{cccccccc}
\toprule
\multicolumn{8}{c}{Emotion class labelling} \\
T   &  Acc   & $\text{NLL}^\text{all}$ & $\text{RMSE}^s$ &$\bar{s}$ & $\mathcal{E}(\Bar{s})$ & $\hat{\kappa}$ & $\mathcal{E}(\hat{\kappa})$  \\
\midrule
0.8 & 0.594 &  1.395  & 0.221
   & 0.200   & 0.044         & 0.307 & 0.053                   \\
0.9 & 0.594 &  1.390  & 0.219
   & 0.216   & 0.029         & 0.259 & 0.005                 \\
1.0   & 0.593 &  1.389  & 0.218
   & 0.229   & 0.015         & 0.222 & 0.032                   \\
1.1 & 0.592 &  1.389  & 0.218
   & 0.241   & 0.004         & 0.191 & 0.063                   \\
1.2 & 0.590 &  1.391  & 0.219   & 0.251   & 0.007         & 0.166 & 0.088      \\
\midrule
\midrule
\multicolumn{8}{c}{Toxic speech detection} \\
T   &  Acc   &  $\text{NLL}^\text{all}$ & $\text{RMSE}^s$ &$\bar{s}$ & $\mathcal{E}(\Bar{s})$ & $\hat{\kappa}$ & $\mathcal{E}(\hat{\kappa})$ \\
\midrule
0.8 & 0.671 &  0.851 & 0.272 & 0.125 & 0.065 & 0.721 & 0.115  \\
0.9 & 0.675 &  0.842 & 0.265 & 0.157 & 0.033 & 0.650 & 0.044  \\
1.0   & 0.673 &  0.837 & 0.263 & 0.188 & 0.002 & 0.580 & 0.026\\
1.1 & 0.671 &  0.836 & 0.264 & 0.216 & 0.026 & 0.512 & 0.094  \\
1.2 & 0.669 &  0.837 & 0.267 & 0.242 & 0.052 & 0.450 & 0.156  \\
\midrule
\midrule
\multicolumn{8}{c}{Speech quality assessment} \\
T   & $\text{RMSE}^{\bar{y}}$ & $\text{NLL}^\text{all}$ & $\text{RMSE}^s$ & $\bar{s}$ & $\mathcal{E}(\Bar{s})$ & $\hat{\kappa}$ & $\mathcal{E}(\hat{\kappa})$  \\ \midrule
0.8 & 0.377     &  1.671    & 0.274           & 0.963   & 0.179          & /&/                                     \\
0.9 & 0.380     & 1.624    & 0.218           & 1.083  & 0.059       & /&/                                    \\
1.0   & 0.384     &  1.611    & 0.223           & 1.201   & 0.059         & /&/                              \\
1.1 & 0.388     &  1.619    & 0.281           & 1.322   & 0.180        & /&/                                 \\
1.2 & 0.393     &  1.640    & 0.371           & 1.440   & 0.299                             & /&/             \\ 
\bottomrule
\end{tabular}
\end{table}

\begin{table}[tb]
\footnotesize
\centering
\caption{Adjusting the diversity of MCDP models by dropout rate}
\label{tab: mcdp diversity}
\begin{tabular}{cccccccc}
\toprule
\multicolumn{8}{c}{Emotion class labelling}\\
dp  &  Acc   & $\text{NLL}^\text{all}$ & $\text{RMSE}^s$ &$\bar{s}$ & $\mathcal{E}(\Bar{s})$ & $\hat{\kappa}$ & $\mathcal{E}(\hat{\kappa})$  \\
\midrule
0.1 & 0.583 & 1.463  & 0.303
   & 0.040 & 0.205         & 0.791 & 0.537                 \\
0.2 & 0.589 & 1.426  & 0.303
   & 0.040 & 0.204         & 0.773 & 0.519                 \\
0.3 & 0.590 & 1.415  & 0.300
   & 0.045 & 0.199         & 0.761 & 0.507              \\
0.4 & 0.585 & 1.405  & 0.296
   & 0.051 & 0.194         & 0.723 & 0.469              \\
0.5 & 0.589 & 1.409  &0.294
   & 0.053 & 0.191         & 0.715 & 0.461            \\
\midrule
\midrule
\multicolumn{8}{c}{Toxic speech detection}\\
dp  &  Acc   & $\text{NLL}^\text{all}$ & $\text{RMSE}^s$ &$\bar{s}$ & $\mathcal{E}(\Bar{s})$ & $\hat{\kappa}$ & $\mathcal{E}(\hat{\kappa})$ \\
\midrule
0.1 & 0.661 & 0.925 & 0.314 & 0.049 & 0.158 & 0.831 & 0.225  \\
0.2 & 0.666 & 0.916 & 0.308 & 0.061 & 0.147 & 0.800 & 0.194\\
0.3 & 0.654 & 0.968 & 0.299 & 0.081 & 0.127 & 0.750 & 0.144  \\
0.4 & 0.662 & 0.943 & 0.297 & 0.085 & 0.122 & 0.731 & 0.125 \\
0.5 & 0.662 & 0.896 & 0.296 & 0.088 & 0.120 & 0.720 & 0.114 \\
\midrule
\midrule
\multicolumn{8}{c}{Speech quality assessment}\\
dp  & $\text{RMSE}^{\bar{y}}$ & $\text{NLL}^\text{all}$ & $\text{RMSE}^s$ & $\bar{s}$ & $\mathcal{E}(\Bar{s})$ & $\hat{\kappa}$ & $\mathcal{E}(\hat{\kappa})$  \\ 
\midrule
0.1 & 0.385     & 1.828    & 0.982     & 0.180   & 0.961     &/&/              \\
0.2 & 0.412     & 1.824    & 0.928     & 0.236   & 0.905       &/&/                 \\
0.3 & 0.408     & 1.797    & 0.871     & 0.294   & 0.847       &/&/                \\
0.4 & 0.367     & 1.805    & 0.938     & 0.227   & 0.915      &/&/                   \\
0.5 & 0.356     & 1.780    & 0.889     & 0.278   & 0.864      &/&/                   \\
\bottomrule
\end{tabular}
\end{table}

\clearpage
\section{Additional Visualization for Emotion Class Labelling}
\label{apdx: cat graph}
\begin{figure}[h]
    \centering
    \includegraphics[width=0.4\textwidth]{figs/legend-grey3.pdf}
    \begin{minipage}[b]{\linewidth}
\centerline{\includegraphics[width=\linewidth]{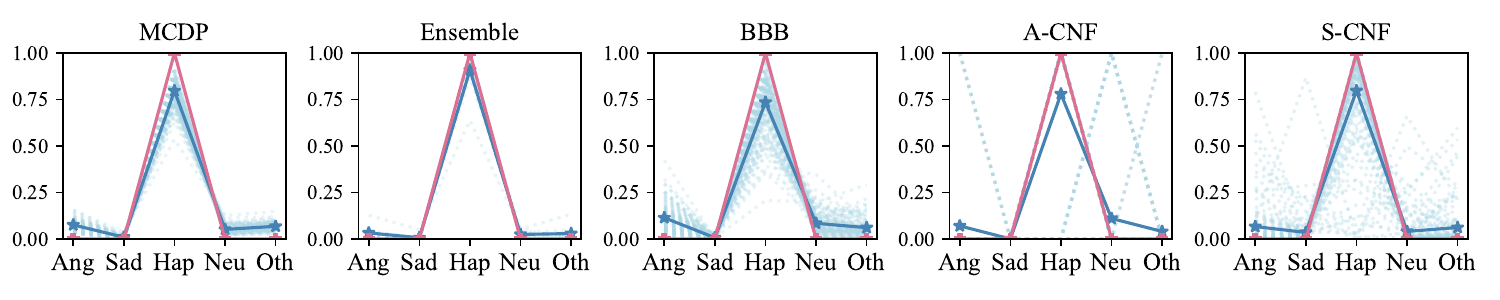}}
\centerline{(a) ``MSP-PODCAST\_1216\_0067.wav''}
    \end{minipage}
    \begin{minipage}[b]{\linewidth}
\centerline{\includegraphics[width=\linewidth]{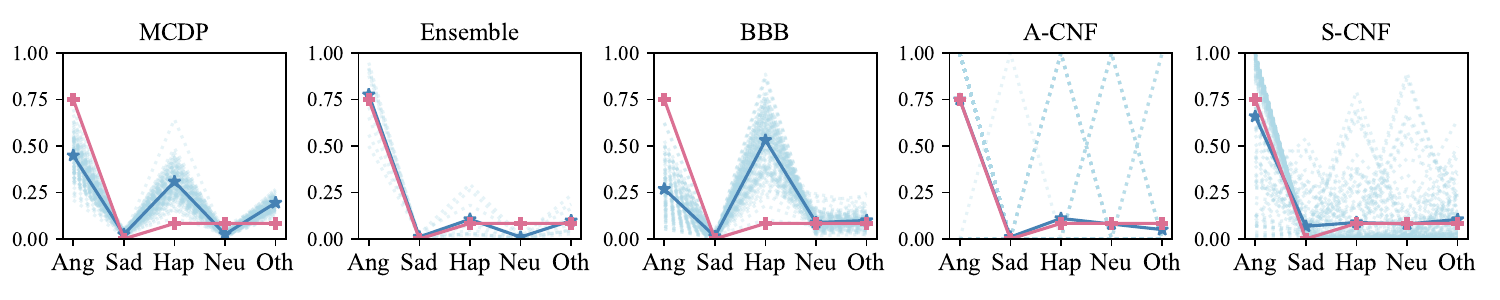}}
\centerline{(b) ``MSP-PODCAST\_0566\_0220''}
    \end{minipage}
    \begin{minipage}[b]{\linewidth}
\centerline{\includegraphics[width=\linewidth]{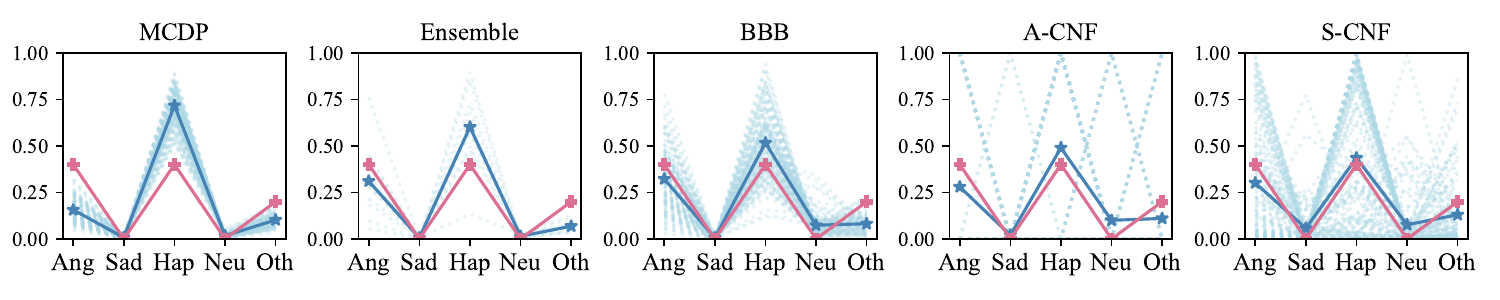}}
\centerline{(c) ``MSP-PODCAST\_0584\_0145.wav''}
    \end{minipage}
    \begin{minipage}[b]{\linewidth}
\centerline{\includegraphics[width=\linewidth]{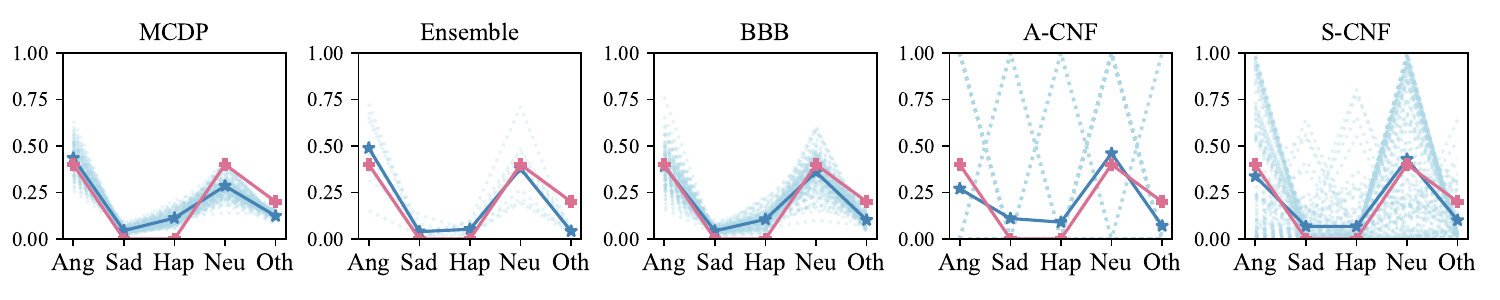}}
\centerline{(d) ``MSP-PODCAST\_0876\_0069.wav''}
    \end{minipage}
    \begin{minipage}[b]{\linewidth}
\centerline{\includegraphics[width=\linewidth]{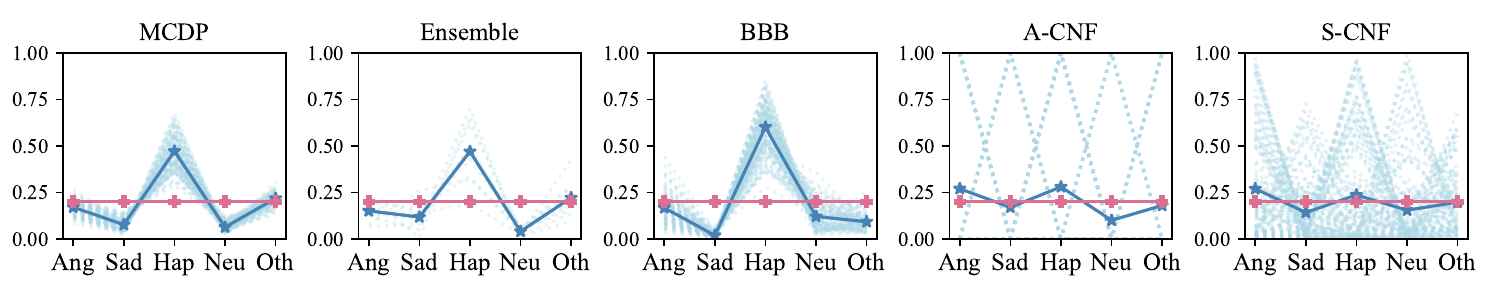}}
\centerline{(e) ``MSP-PODCAST\_0587\_0073.wav''}
    \end{minipage}
    
    \caption{Additional visualized examples for emotion class labelling. The y-axis corresponds to the probability mass. Each sample is a categorical distribution.
    The probability mass values of different categories in each categorical distribution are connected for the purpose of better visualization. CVAE is omitted because it collapses to one category for all inputs.}
    \label{fig: more emo visual}
\end{figure}

This section shows additional visualized examples for emotion class labelling when human annotators reach a consensus (Figure~\ref{fig: more emo visual}~(a)(b)), diverge (Figure~\ref{fig: more emo visual}~(c)(d)), and give distinct labels (Figure~\ref{fig: more emo visual}~(e)). The proposed S-CNF can better simulate the aggregated behaviour as well as the variability of human annotations in all cases.

\section{Additional Visualization for Toxic Speech Detection}
\begin{figure}[h]
    \centering
    \includegraphics[width=0.4\textwidth]{figs/legend-grey3.pdf}
    \begin{minipage}[b]{\linewidth}
\centerline{\includegraphics[width=\linewidth]{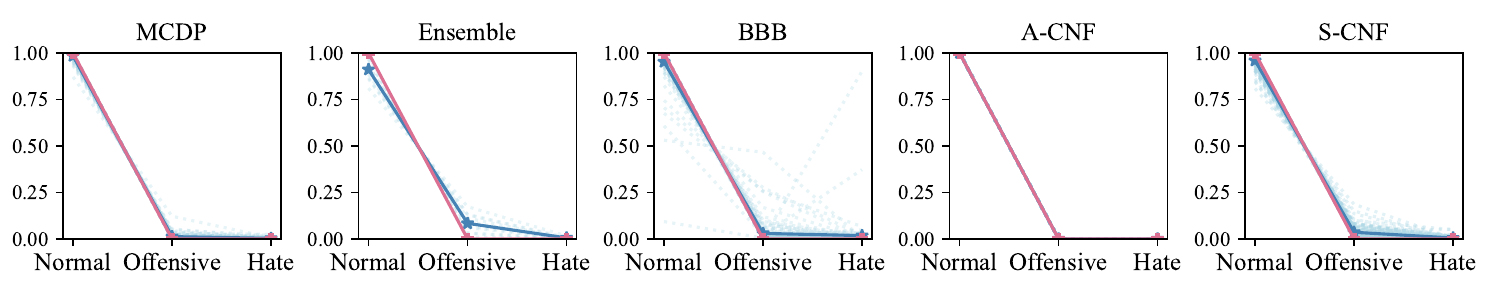}}
\centerline{(a) ``1092591391086178304\_twitter''}
    \end{minipage}
    \begin{minipage}[b]{\linewidth}
\centerline{\includegraphics[width=\linewidth]{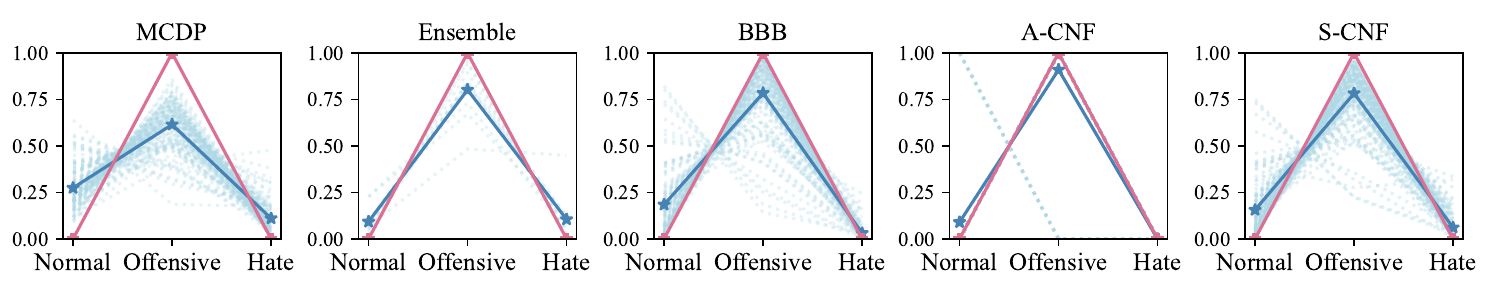}}
\centerline{(b) ``10665227\_gab''}
    \end{minipage}
    \begin{minipage}[b]{\linewidth}
\centerline{\includegraphics[width=\linewidth]{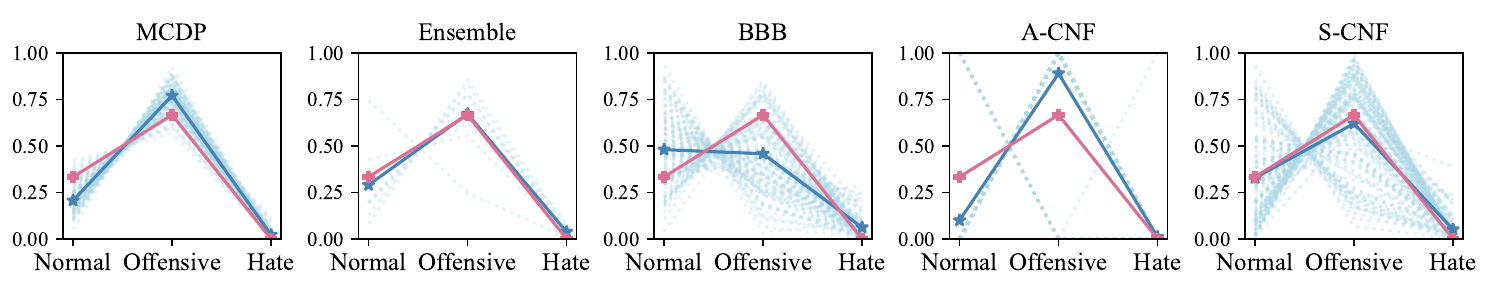}}
\centerline{(c) ``1179093526098743296\_twitter''}
    \end{minipage}
    \begin{minipage}[b]{\linewidth}
\centerline{\includegraphics[width=\linewidth]{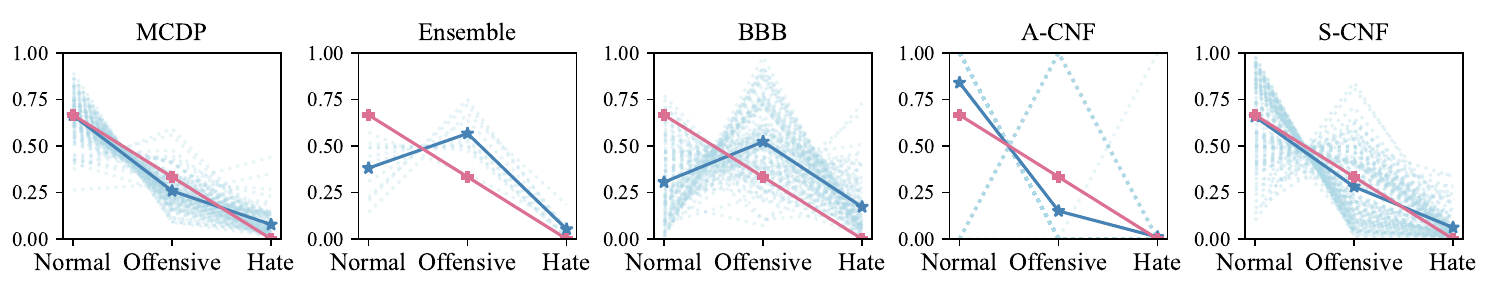}}
\centerline{(d) ``18777413\_gab''}
    \end{minipage}
    \begin{minipage}[b]{\linewidth}
\centerline{\includegraphics[width=\linewidth]{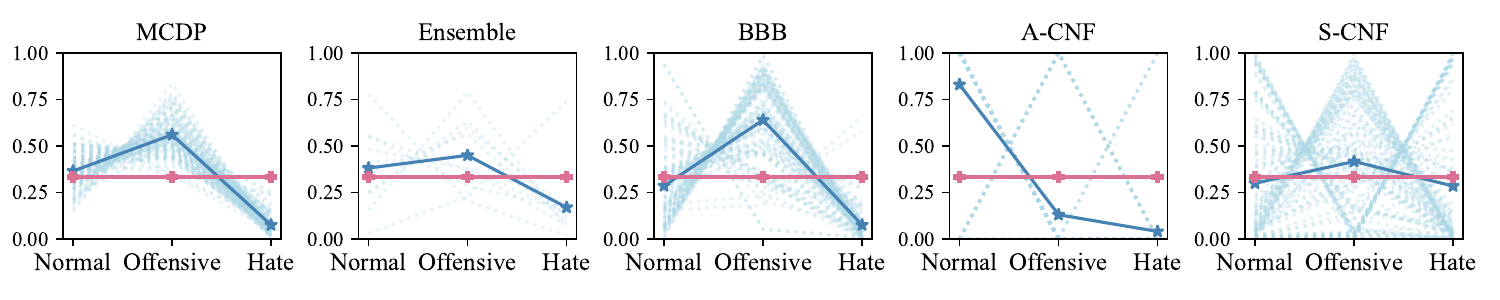}}
\centerline{(e) ``20362058\_gab''}
    \end{minipage}
    \caption{Additional visualized examples for toxic speech detection. The y-axis corresponds to the probability mass. Each sample is a categorical distribution.
    The probability mass values of different categories in each categorical distribution are connected for the purpose of better visualization. CVAE is omitted because it collapses to one category for all inputs.}
    \label{fig: more hate visual}
\end{figure}

This section shows visualized examples for toxic speech detection when all three human annotators provide the same label (Figure~\ref{fig: more hate visual}~(a)(b)), one of them gives a different label (Figure~\ref{fig: more hate visual}~(c)(d)), and all three annoators give distinct labels (Figure~\ref{fig: more hate visual}~(e)). The proposed S-CNF can better simulate the aggregated behaviour as well as the variability of human annotations in all cases.
\label{apdx: hate graph}

\section{Additional Visualization for Speech Quality Evaluation}
\label{apdx: reg graph}
This section presents several additional visualized cases for speech quality evaluation. Generated samples (before rounding) are plotted in the sub-figures on the left. For clearer visualization, the samples are spread along $y$ axis according to density to avoid overlapping. As can be seen, samples generated by the proposed I-CNF method (in blue) can better simulate the diversity of human annotations (in pink).

\begin{figure}[H]
    \centering
    \includegraphics[width=0.95\linewidth]{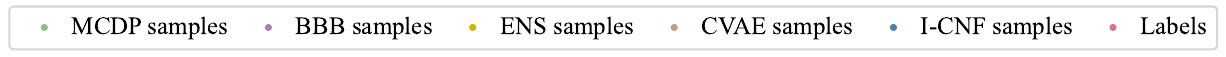}
    \begin{minipage}[b]{0.49\linewidth}
\centerline{\includegraphics[width=\linewidth]{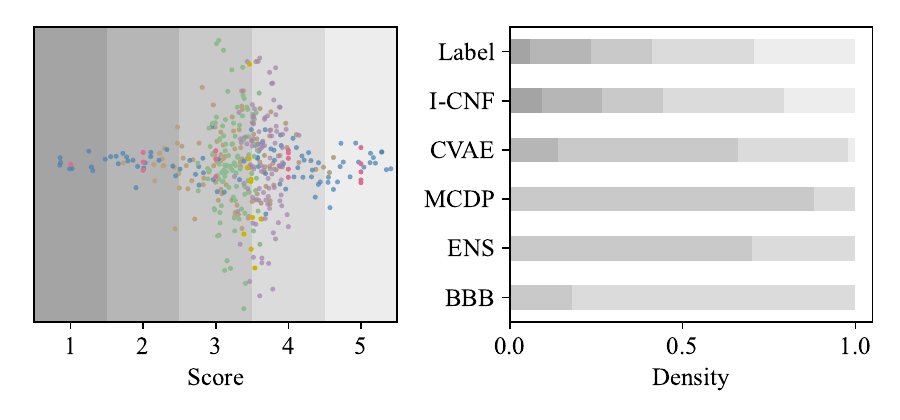}}
\centerline{(a) ``conv\_2007\_0090\_057''}
    \end{minipage}
    \begin{minipage}[b]{0.49\linewidth}
\centerline{\includegraphics[width=\linewidth]{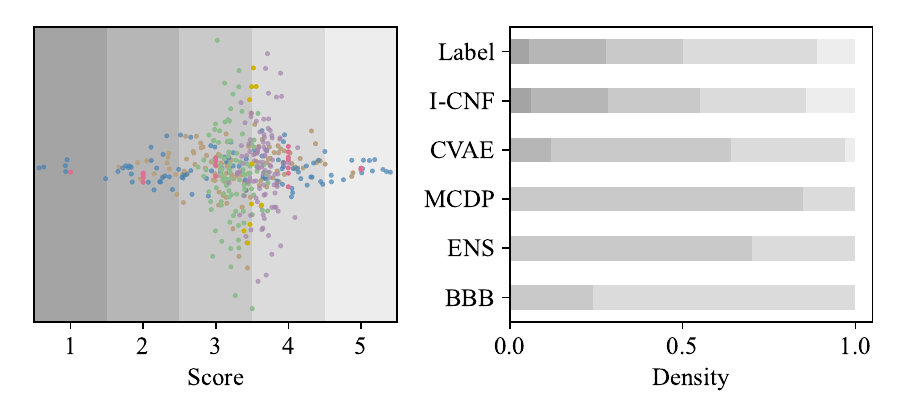}}
\centerline{(b) ``conv\_2007\_0090\_060''}
    \end{minipage}
    \begin{minipage}[b]{0.49\linewidth}
\centerline{\includegraphics[width=\linewidth]{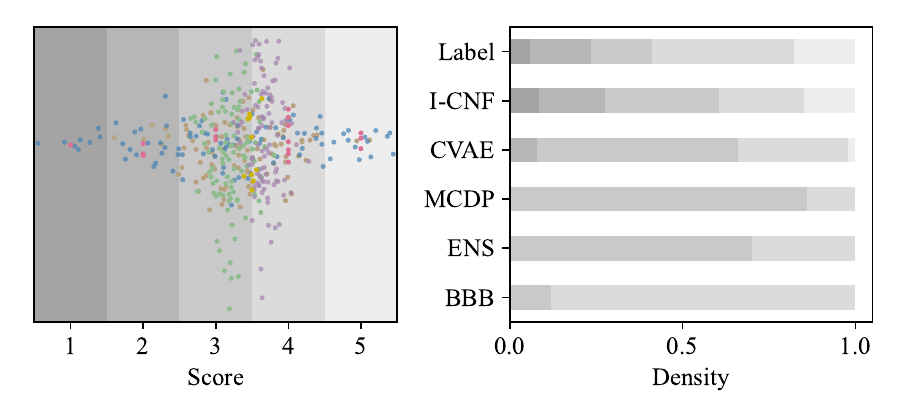}}
\centerline{(c) ``reportorial\_2011\_0154\_145''}
    \end{minipage}
    \begin{minipage}[b]{0.49\linewidth}
\centerline{\includegraphics[width=\linewidth]{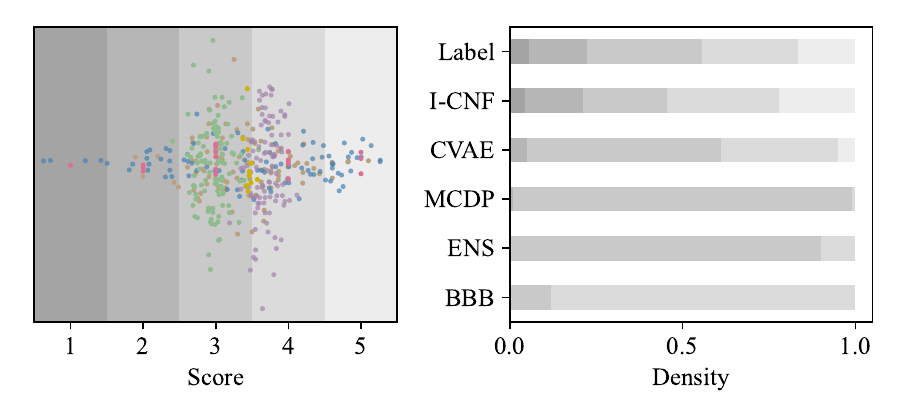}}
\centerline{(d) ``reportorial\_2011\_0096\_186''}
    \end{minipage}
    \begin{minipage}[b]{0.49\linewidth}
\centerline{\includegraphics[width=\linewidth]{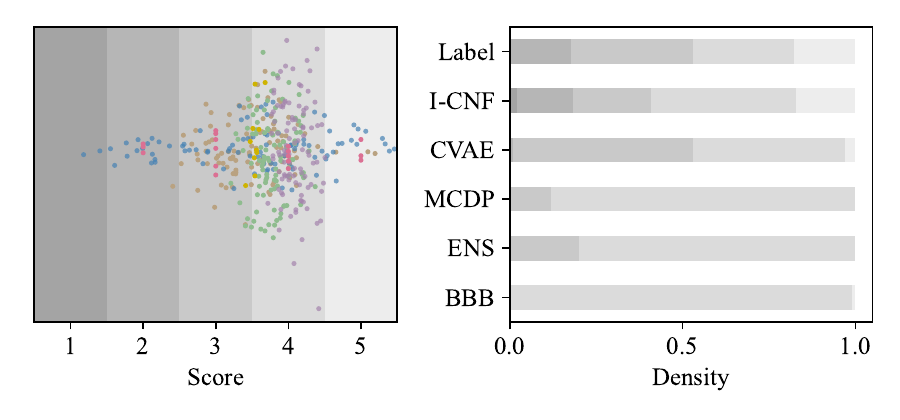}}
\centerline{(e) ``novel\_2011\_0011\_110''}
    \end{minipage}
    \begin{minipage}[b]{0.49\linewidth}
\centerline{\includegraphics[width=\linewidth]{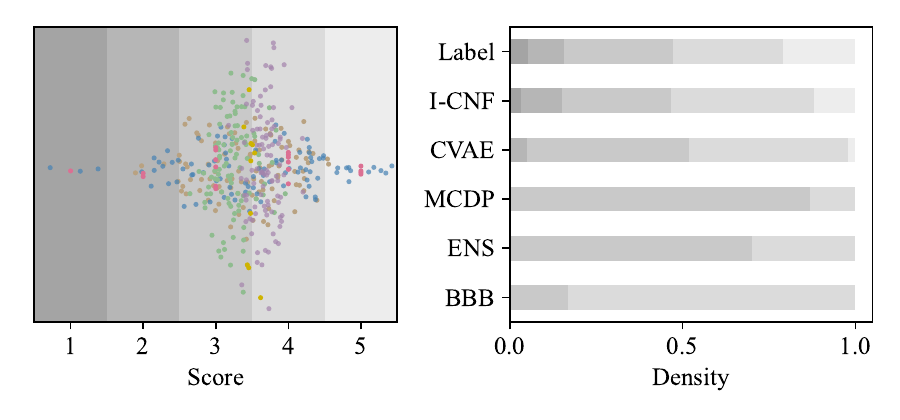}}
\centerline{(f) ``wiki\_0033\_055''}
    \end{minipage}
        \caption{Additional visualized examples for speech quality assessment. For the visualization purpose, the points that have same x values are spread along y axis according to density to avoid overlapping.}
    \label{fig: More MOS visual}
\end{figure}

\end{document}